\documentclass[10pt,twocolumn,letterpaper]{article}
\usepackage[pagenumbers]{cvpr} 

\usepackage[T1]{fontenc}
\usepackage{graphicx}
\usepackage{xcolor}
\usepackage{amsmath,amssymb,amsthm}
\usepackage{booktabs}
\usepackage{multirow}
\usepackage{microtype}
\usepackage{algorithm}
\usepackage{algpseudocode}
\usepackage{natbib}

\newcommand{\best}[1]{\textbf{#1}}
\newcommand{\second}[1]{\underline{#1}}

\definecolor{metricgreen}{RGB}{0,115,65}
\definecolor{metricred}{RGB}{185,35,35}
\newcommand{\metricgain}[1]{\textcolor{metricgreen}{\scriptsize$(+#1)$}}
\newcommand{\metricloss}[1]{\textcolor{metricred}{\scriptsize$(-#1)$}}

\newcommand{\dlow}{\delta_{\mathrm{g}}}
\newcommand{\dhigh}{\delta_{\mathrm{r}}}

\DeclareMathOperator{\TV}{TV}

\graphicspath{{figures/}}

\theoremstyle{plain}

\ifdefined\theorem
\else
  \newtheorem{theorem}{Theorem}
\fi

\ifdefined\lemma
\else
  \newtheorem{lemma}[theorem]{Lemma}
\fi

\ifdefined\proposition
\else
  \newtheorem{proposition}[theorem]{Proposition}
\fi

\ifdefined\corollary
\else
  \newtheorem{corollary}[theorem]{Corollary}
\fi

\theoremstyle{remark}

\ifdefined\remark
\else
  \newtheorem*{remark}{Remark}
\fi

\theoremstyle{plain}

\definecolor{cvprblue}{rgb}{0.21,0.49,0.74}
\usepackage[pagebackref,breaklinks,colorlinks,allcolors=cvprblue]{hyperref}
\usepackage[capitalize]{cleveref}

\def\confName{CVPR}
\def\confYear{2027}

\title{DUNE: SVD-Refined Dual-Resolution Unified Adversarial Noise for Effective Face-Swap Protection}

\author{Sungwon Cho\thanks{Equal contribution.} \qquad Kwanghyun Ko\footnotemark[1] \qquad Jaehyoung Jeon\footnotemark[1] \qquad Myungjoo Kang\thanks{Corresponding author.}\\
Seoul National University}

\usepackage{float}\usepackage{capt-of}
\usepackage{array}
\usepackage{placeins}

\begin{document}
\maketitle
\begin{abstract}
Proactive face-swap protection must disrupt identity transfer while preserving portrait quality. We observe that later singular components of PGD perturbations are more high-frequency, and together they carry most of the perturbation's high-frequency energy. Such high-frequency noise degrades image quality, motivating selective refinement during optimization. We propose DUNE, which jointly optimizes an SVD-refined full-resolution perturbation and a bilinearly upsampled low-resolution perturbation through frozen surrogate recognizers under a shared \(\ell_\infty\) bound. SVD truncation and Gaussian smoothing(SVDG) reduce distortion, while the auxiliary branch recovers protection. Importance masking and masked total variation further regulate local appearance. Across four black-box face swappers, DUNE achieves mean protection rates of 95.1\% on CelebA-HQ and 93.1\% on VGGFace2, leading the compared methods while attaining the best source PSNR, SSIM, and LPIPS. Component-wise ablations and comparisons further show how each component contributes to these gains in protection and image fidelity.
\end{abstract}

\section{Introduction}
\label{sec:intro}

Deepfake techniques such as face swapping can use publicly shared portraits to create realistic images of a person without their permission~\cite{simswap,uniface,faceadapter}. This can pose serious privacy risks and lead to social harm. To address these risks, various facial privacy methods have been developed to disrupt identity recognition while preserving the original appearance of shared portraits. Diffusion-based methods~\cite{diffprotect,diffpurify} use generative priors, but repeated generation incurs high computational costs and may alter source details. Makeup-based methods~\cite{amtgan,clip2protect,diffam} integrate protection into edits that may introduce unwanted makeup or style changes. Noise-based methods instead optimize small perturbations added to the image, aiming to disrupt identity recognition without visible changes~\cite{pgd,tipim,tidim,mifgsm}. This makes them a natural choice when protection and source preservation are both important.

\begin{figure}[!t]
\centering
\includegraphics[width=\columnwidth]{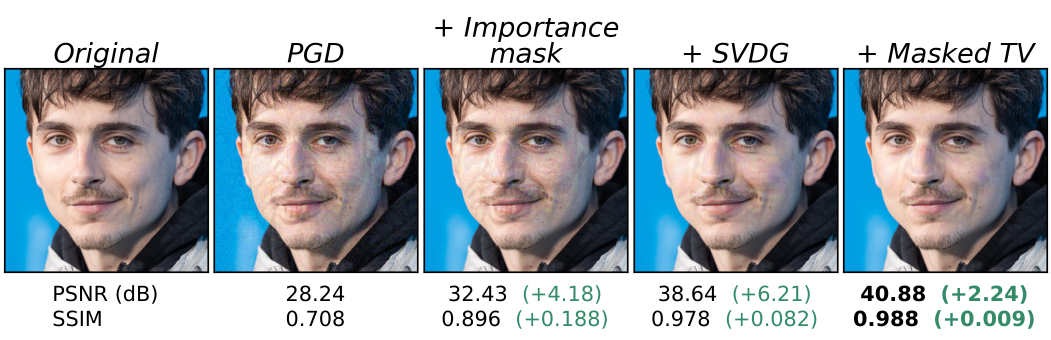}
\caption{\textbf{Visual quality of protected portraits.} High-frequency noise in PGD leaves visible artifacts and degrades image quality. Our importance masking, SVD--Gaussian refinement (SVDG), and masked TV progressively reduce these artifacts. Portrait from~\cite{krichel2025portrait}.}
\label{fig:intro-refinement}
\end{figure}
However, an $\ell_\infty$ constraint limits the magnitude of adversarial perturbations but does not control their spatial structure.
Consequently, high-frequency variations may introduce visible artifacts and degrade image quality~\cite{zhang2020smooth}, as illustrated by the projected gradient descent (PGD)~\cite{pgd} result in \Cref{fig:intro-refinement}.
Although smoothing and fixed-frequency filtering can suppress
such variations~\cite{jeong2025faceshield}, they may also remove
perturbation components useful for protection.
This motivates an instance-adaptive refinement strategy that selectively preserves dominant perturbation structures while reducing distortion.

To investigate this possibility, we decompose PGD perturbations using singular value decomposition (SVD), which selects dominant components in a perturbation-dependent basis rather than a fixed frequency basis.
We observe that later components contain a higher proportion of high-frequency energy (\Cref{fig:component}).
Controlled rank experiments further show that selective SVD truncation during optimization can improve recognition-dodging PSR and source fidelity simultaneously (\Cref{tab:svd-only}).
We calibrate the truncation rank offline using the Gavish--Donoho (GD) hard threshold ~\citep{gavish2014optimal}, providing a data-driven criterion for rank selection.

Based on these findings, we propose DUNE, a dual-resolution pipeline for face-swap protection. Its main full-resolution branch applies channelwise top-$r$ SVD truncation followed by Gaussian smoothing during optimization. Together, these operations form the SVD--Gaussian refinement (SVDG) block.
While SVDG adaptively refines dominant singular components across the full-resolution perturbation, we introduce an auxiliary low-resolution branch to provide complementary spatial adjustments.
Bilinear upsampling distributes each grid value over a local neighborhood, enabling localized corrections with low-rank structure and a low-frequency bias.
Both branches are jointly optimized through
frozen surrogate recognizers under a shared
perturbation budget.
We also build an importance mask to reduce unnecessary noise and focus perturbations on facial regions important for identity recognition. The mask is computed from identity-loss gradients and applied to both branches, while a masked total variation (TV) loss reduces sharp changes in the perturbation within the selected regions.

On CelebA-HQ and VGGFace2, DUNE achieves the highest protection rates among the compared methods across all four evaluated face swappers, together with the best source PSNR, SSIM, and LPIPS in the main comparison. Our contributions are threefold:
\begin{itemize}
    \item We identify frequency trends across the singular components of adversarial perturbations and show through controlled rank experiments that SVD truncation alone can improve both protection and source fidelity.
    \item We propose DUNE, a dual-resolution pipeline that combines SVD refinement and bilinear upsampling to regulate high-frequency variation, with importance masking and masked total variation providing local appearance control.
    \item We validate the contribution of each component through ablation studies and assess the protection--fidelity balance through budget-matched recognition dodging and black-box face-swap comparisons.
\end{itemize}

\section{Related Work}
\label{sec:related}

\noindent\textbf{Image Protection Methods.}
Facial privacy methods protect shared portraits through additive noise, adversarial makeup, or diffusion-based generation. Noise-based approaches optimize bounded perturbations~\cite{pgd}, accumulating gradients~\cite{mifgsm} or averaging over translations~\cite{tidim} and transformed inputs~\cite{dim} for transferability, with identity objectives disrupting recognition~\cite{tipim}. Makeup-based approaches transfer reference appearance~\cite{amtgan}, optimize generator latents using text prompts~\cite{clip2protect}, or combine makeup transfer with diffusion-based identity disruption~\cite{diffam}. Diffusion-based approaches optimize diffusion-autoencoder latents~\cite{diffprotect} or unconditional embeddings and adversarial codes to resist purification~\cite{diffpurify}. For face-swap protection, NullSwap~\cite{wang2025nullswap}
learns identity-guided perturbations to cloak source identities
in a black-box setting. FaceShield~\cite{jeong2025faceshield}
targets diffusion attention and facial feature extractors,
using Gaussian blur and low-pass filtering to improve
imperceptibility and JPEG robustness. DUNE jointly optimizes
an SVD-refined full-resolution perturbation and a bilinearly
upsampled auxiliary perturbation under a shared $\ell_\infty$
bound to balance protection and source fidelity.

\noindent\textbf{Face Swapping.}
Face swapping transfers source identity while preserving target attributes. Existing approaches combine identity injection with weak feature matching~\cite{simswap}, unify reenactment and swapping~\cite{uniface}, or control identity, structure, and attributes in pretrained diffusion models~\cite{faceadapter}. We evaluate these approaches together with a publicly available face-swapping model~\cite{inswapper}, using held-out recognizers to assess protection against downstream identity transfer.

\noindent\textbf{SVD and Spectral Refinement.}
Truncated SVD provides a best low-rank approximation~\cite{eckartyoung1936} in a matrix-dependent basis, unlike the fixed cosine basis of DCT~\cite{ahmed1974dct}. Adversarial attacks also use nuclear-norm constraints and low-rank factorization~\cite{kazemi2021structured,lowrankpgd2024}.
Prior low-rank attacks optimize factorized perturbations for classification and adversarial training~\cite{lowrankpgd2024}. DUNE instead refines a full-resolution identity-disrupting perturbation after each update and jointly optimizes a bilinearly upsampled branch under a shared \(\ell_\infty\) bound.
The Marchenko--Pastur law~\cite{marchenko1967} characterizes noise spectra, and Gavish--Donoho hard thresholding~\cite{gavish2014optimal} guides singular-value selection for matrix recovery. These results provide references for rank calibration rather than guarantees for adversarial protection.

\section{Method}
\label{sec:method}

\subsection{Analysis of Perturbation Components}
\label{sec:component-analysis}
\begin{figure}[!t]
\centering
\includegraphics[width=\columnwidth]{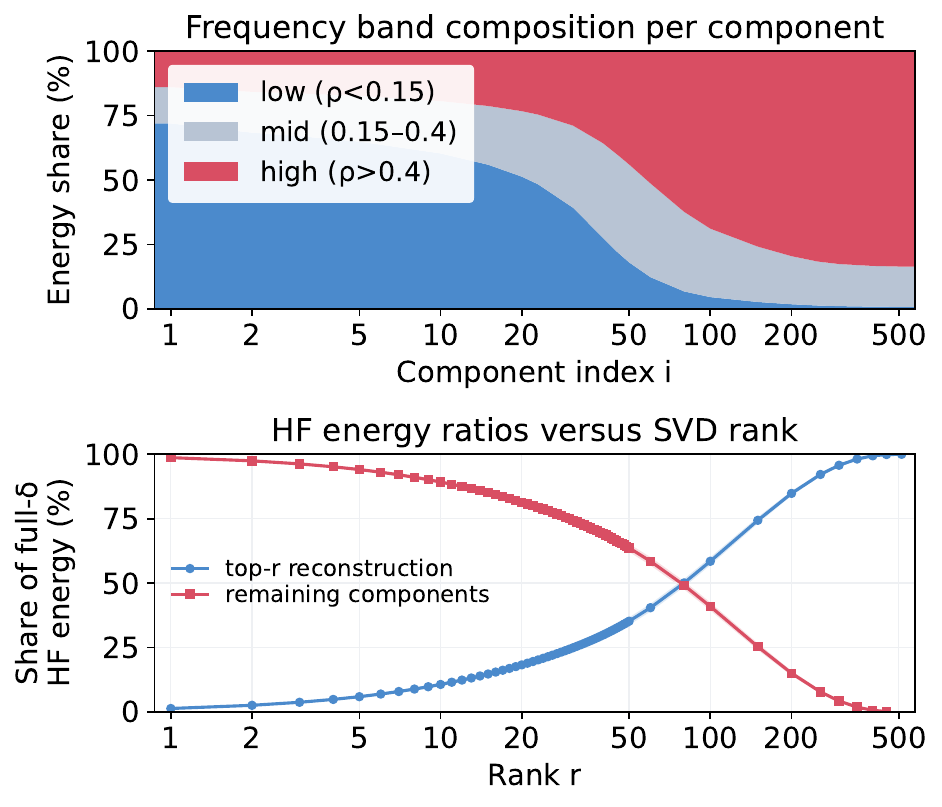}
\caption{\textbf{Singular-component frequency profiles.} PGD perturbations on 1{,}000 CelebA-HQ images at $512\times512$. Top: frequency-band energy shares per singular component. Bottom: HF energy ($\rho>0.4$) of the top-$r$ reconstruction and the remaining components, normalized by the full perturbation's HF energy and averaged over images.}
\label{fig:component}
\end{figure}

We analyze PGD perturbations~\cite{pgd} on 1{,}000 CelebA-HQ
images at $512\times512$ resolution using channelwise SVD.
We order the resulting rank-one components by decreasing singular
value and examine their frequency profiles. This analysis reveals
that later components have lower individual energy but a higher
proportion of high-frequency content (\Cref{fig:component}, upper panel).

To examine how later components affect the full perturbation,
we combine them into a residual and measure its HF energy
($\rho>0.4$), along with that of the top-$r$
reconstruction, relative to the full perturbation's HF energy.
At rank 11, the residual-to-full ratio of high-frequency energy is 88.4\% on average (\Cref{fig:component}, bottom).
\Cref{tab:svd-only} examines how rank constraints during
optimization affect protection, image fidelity, and HF energy
share on 500 CelebA-HQ pairs at $256\times256$ resolution.
We vary the number of leading singular components retained
per channel after each gradient update and compare the
results with an untruncated baseline.
\begin{table}[!t]
\centering\small
\caption{\textbf{SVD truncation.} CelebA-HQ, 500 pairs. PSR is the dodging PSR (\%) averaged over four held-out recognizers. Full-resolution perturbations use $\epsilon=16/255$, 100 gradient-direction steps, $\alpha=1/255$. HF is the percentage of each optimized perturbation's total energy at normalized radial frequency $\rho>0.4$.}
\label{tab:svd-only}
\setlength{\tabcolsep}{3.5pt}
\begin{tabular}{lcccccc}
\toprule
Rank $r$ & PSR$\uparrow$ & PSNR$\uparrow$ & SSIM$\uparrow$ & FID$\downarrow$ & LPIPS$\downarrow$ & HF$\downarrow$ \\
\midrule
None & 86.10 & 27.78 & 0.652 & 70.2 & 0.2160 & 52.6 \\
11 & 85.50 & \best{31.47} & \best{0.854} & \best{45.3} & \best{0.0975} & \best{21.7} \\
20 & \best{88.25} & 30.95 & 0.827 & 51.2 & 0.1130 & 25.3 \\
30 & 87.75 & 30.50 & 0.803 & 56.0 & 0.1269 & 29.2 \\
45 & 87.90 & 29.99 & 0.774 & 59.9 & 0.1434 & 34.3 \\
\bottomrule
\end{tabular}
\end{table}
\Cref{tab:svd-only} shows that retaining 20 singular components per channel increases recognition-dodging PSR from 86.10\% to 88.25\% and PSNR from 27.78 to 30.95 dB. The rank of 11 used in DUNE increases PSNR to 31.47 dB, with a 0.60-percentage-point decrease in dodging PSR relative to the untruncated baseline.

Together with \Cref{fig:component}, which shows that later singular components collectively carry substantial HF energy, these results support selective truncation as a means of improving the protection–fidelity balance. We therefore apply SVD refinement after each gradient update, allowing subsequent updates to adapt to this restriction.

\begin{figure*}[!t]
\centering
\includegraphics[width=\textwidth]{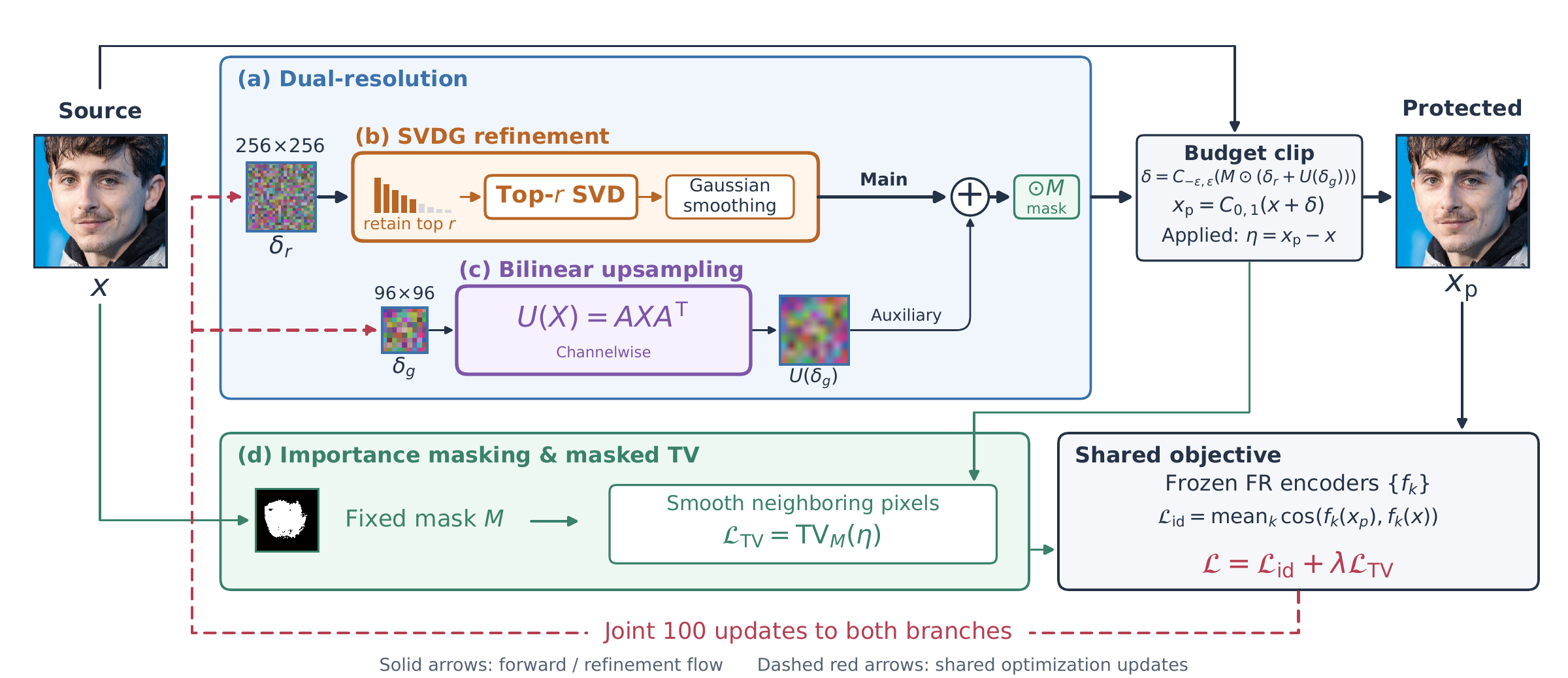}
\caption{\textbf{Overview of DUNE.} The main full-resolution branch applies SVDG (b); an auxiliary low-resolution perturbation is bilinearly upsampled (c). A single importance mask (d) is applied to their sum, which is then clipped to the perturbation budget and image range. Both branches share an identity-suppression objective through frozen recognizers, with masked total variation regularizing the applied perturbation $\eta$. Portrait from~\cite{krichel2025portrait}.}
\label{fig:pipeline}
\end{figure*}

\subsection{Overview and Objective}
\label{sec:formulation}
Given a portrait $x\in[0,1]^{3\times H\times W}$, we produce a protected source $x_{\mathrm p}$ satisfying $\|x_{\mathrm p}-x\|_\infty\le\epsilon$.
As shown in~\Cref{fig:pipeline}(a), DUNE jointly optimizes
a main full-resolution perturbation $\dhigh$ and an auxiliary
low-resolution grid $\dlow$. The main branch applies SVDG refinement
after each gradient update (b), while the auxiliary grid
is bilinearly upsampled by $U$ (c).
A single fixed binary importance mask $M$ (d) is applied
after summing the two branch outputs.
Both branches share an objective through frozen surrogate
recognizers $\{f_k\}_{k=1}^{K}$, without access to the
downstream swapper or held-out evaluator.
For $a\le b$, define elementwise clipping as $C_{a,b}(z)=\min\{b,\max\{a,z\}\}$.
We form
\begin{align}
\delta&=C_{-\epsilon,\epsilon}\!\left(M\odot\big(\dhigh+U(\dlow)\big)\right),\label{eq:delta}\\
x_{\mathrm p}&=C_{0,1}(x+\delta),\qquad \eta=x_{\mathrm p}-x.\nonumber
\end{align}
Here $M$ is defined in \Cref{sec:mtv}, $\delta$ is the masked and budget-clipped sum, and $\eta$ is the applied perturbation. Image-range clipping cannot increase the deviation from $x$, so $\|\eta\|_\infty\le\epsilon$.
We jointly minimize
\begin{equation}
\mathcal L=\frac{1}{K}\sum_{k=1}^{K}
\cos\!\left(f_k(x_{\mathrm p}),f_k(x)\right)
+\lambda\TV_M(\eta).
\label{eq:loss}
\end{equation}
Here $\cos$ denotes cosine similarity and $\lambda\ge0$ weights masked TV, defined in \Cref{sec:mtv}. Both branches obey individual
budgets $\epsilon_r$ and $\epsilon_g$ in addition
to the global budget $\epsilon$.

\subsection{SVD-Refined Full-Resolution Perturbation}
\label{sec:svd}

The full-resolution main branch optimizes a perturbation
$\dhigh\in\mathbb{R}^{3\times H\times W}$ using
an SVD--Gaussian (SVDG) refinement block.
The block applies channelwise top-$r$ SVD truncation
to retain dominant singular components, followed by
Gaussian smoothing to reduce local spatial variation.
We compare SVD truncation, Gaussian smoothing, and their combination in \Cref{tab:refinement}.
Since SVD refinement is applied independently to each
color channel, we consider a single channel and omit
the channel index for simplicity.
Let $\delta_r^{\mathrm{pre}}\in\mathbb{R}^{H\times W}$
denote the perturbation channel immediately before
SVD truncation.
With singular values
$s_1\ge\cdots\ge s_{\min(H,W)}$, we define
\begin{equation}
  \mathcal P_r(\delta_r^{\mathrm{pre}})
  =\sum_{i=1}^{r}s_i u_i v_i^\top.
  \label{eq:svd}
\end{equation}
The retained singular vectors are determined by
the current perturbation rather than a fixed
spatial or frequency basis.
Since Gaussian smoothing is a separable linear
operation, it does not increase matrix rank
when applied alone.
The rank constraint applies to the SVD-truncated
matrix; other subsequent operations may alter it.

\noindent\textbf{Selection of the truncation rank.}
We determine $r$ using the Gavish--Donoho (GD) hard threshold for an unknown noise level ~\citep{gavish2014optimal}.
For a square perturbation channel, the threshold is $\widehat{\tau}_{\mathrm{GD}} \approx 2.858\,s_{\mathrm{med}}$, where $s_{\mathrm{med}}$ is the median singular value.
We apply this criterion channelwise to PGD perturbations generated at \(256\times256\) on CelebA-HQ images not used for evaluation.
This yields a median retained rank of 11,
which we use as the fixed truncation rank for DUNE.

\noindent\textbf{When does SVD remove high-frequency content?}
The frequency profiles in \Cref{fig:component}
provide empirical motivation for retaining leading
singular components. We next give a sufficient condition
under which this selection removes high-frequency (HF)
content.

Consider an idealized pre-truncation perturbation channel
\begin{equation}
  \delta_r^{\mathrm{pre}}=S+E,
  \qquad E=P_HGP_H,
\end{equation}
where $S$ is a rank-$r$ low-frequency component,
$P_H$ is the orthogonal projector onto an
$m$-dimensional high-frequency subspace, and
$G_{ij}$ are independently sampled from
$\mathcal{N}(0,\sigma^2/m)$.

\par
\begingroup
\ifdim\dimexpr\pagegoal-\pagetotal\relax<10\baselineskip
  \newpage
\fi
\endgroup
\begin{proposition}[Removal of Gaussian HF Components]
\label{prop:main-svd-hf}
Suppose $\delta_r^{\mathrm{pre}}=S+E$, where
$\operatorname{rank}(S)=r$,
$S=(I-P_H)S(I-P_H)$, and $E=P_HGP_H$.
If
\begin{equation}
  s_r(S)>\|E\|_2,
\end{equation}
then the rank-$r$ truncated SVD exactly recovers
the low-frequency component:
\begin{equation}
  \mathcal P_r(\delta_r^{\mathrm{pre}})=S.
  \label{eq:main-svd-hf-recovery}
\end{equation}
\end{proposition}

Under this Gaussian model, $\|E\|_2$ approaches
$2\sigma$ as $m$ increases. Thus, sufficiently
dominant low-frequency components can be preserved
while the Gaussian HF component is removed by
SVD truncation.

This result illustrates a sufficient condition rather
than a general guarantee for PGD perturbations.
The practical effect of SVD truncation alone is
evaluated empirically in~\Cref{tab:svd-only}.
The proof is provided in the appendix.
The comparison with frequency filtering and SVDG
is reported in~\Cref{tab:refinement}.

\subsection{Bilinearly Upsampled Low-Resolution Perturbation}
\label{sec:lowres}

We optimize a low-resolution perturbation
$\dlow\in\mathbb{R}^{3\times h\times w}$,
where $h<H$ and $w<W$, and upsample it using
bilinear interpolation $U$.

Since interpolation is applied independently to each
color channel, consider a single channel
$\delta_g^{(c)}\in\mathbb{R}^{N\times N}$.
For a square image, let
$A\in\mathbb{R}^{M\times N}$ denote the
one-dimensional interpolation matrix.
The upsampled channel satisfies
\begin{equation}
    U(\delta_g^{(c)})
    =A\delta_g^{(c)}A^\top.
\end{equation}
Since $A$ has full column rank,
\begin{equation}
    \operatorname{rank}(U(\delta_g^{(c)}))
    =\operatorname{rank}(\delta_g^{(c)})\le N.
    \label{eq:main-grid-rank}
\end{equation}
Thus, upsampling a $96\times96$ perturbation
to $256\times256$ restricts the output rank
to at most $96$.

\noindent\textbf{Frequency concentration.}
Bilinear interpolation constrains the frequency
content of the upsampled perturbation.
Using the normalized radial FFT frequency in
\Cref{fig:component}, we define HF energy
by $\rho>0.4$.
For $96\times96$ to $256\times256$ half-pixel
bilinear upsampling, the HF energy fraction
is bounded above by $32.38\%$ for any nonzero
real-valued input grid.
This sharp deterministic bound applies before
masking, branch combination, and clipping.
The derivation and interpolation convention
are provided in the appendix.

\subsection{Dual-Resolution Perturbations}
\label{sec:dual}

At the SVD stage, the full-resolution branch retains
an adaptive rank-$r$ representation, whereas bilinear
upsampling provides a fixed interpolation space with
rank up to $N$.
The following proposition characterizes how these
distinct constraints complement each other.

\begin{proposition}[Complementary Representation Spaces]
\label{prop:main-dual-expressivity}
Let $A\in\mathbb{R}^{M\times N}$ have full column rank,
with $1\le r<N<M$.
Define the representation classes
$\mathcal I_A=\{AZA^\top:Z\in\mathbb{R}^{N\times N}\}$
and
$\mathcal R_r=\{R\in\mathbb{R}^{M\times M}:
\operatorname{rank}(R)\le r\}$.
Neither class contains the other, and their sum
strictly extends both.
Moreover, for $X\in\mathcal I_A$ and
$Y\in\mathcal R_r$,
\begin{equation}
    \operatorname{rank}(X+Y)
    \le\min\{M,N+r\}.
    \label{eq:main-dual-rank}
\end{equation}
\end{proposition}

The SVD branch can represent low-rank directions
outside the interpolation space, while the
low-resolution branch adds structures beyond
the rank-$r$ representation.
Their joint optimization provides additional
flexibility to recover protection under the
appearance constraints, as evaluated in
\Cref{sec:ablation-studies}.
The proposition applies before subsequent
processing; its proof is given in the
appendix.

\subsection{Importance-Guided Masked TV}
\label{sec:mtv}
\begin{table*}[!t]
\centering\small
\caption{\textbf{Source fidelity and black-box recognition dodging.} PSR/R1-P/R5-P (\%) average four held-out recognizers. Bold/underline: best/second best, including ties.}
\label{tab:dodging}
\setlength{\tabcolsep}{3pt}
\resizebox{\textwidth}{!}{%
\begin{tabular}{llccccccc|ccccccc}
\toprule
& & \multicolumn{7}{c|}{CelebA-HQ} & \multicolumn{7}{c}{VGGFace2} \\
Type & Method & PSR$\uparrow$ & R1-P$\uparrow$ & R5-P$\uparrow$ & PSNR$\uparrow$ & SSIM$\uparrow$ & FID$\downarrow$ & LPIPS$\downarrow$ & PSR$\uparrow$ & R1-P$\uparrow$ & R5-P$\uparrow$ & PSNR$\uparrow$ & SSIM$\uparrow$ & FID$\downarrow$ & LPIPS$\downarrow$ \\
\midrule
 & Clean & 9.00 & 11.95 & 8.40 & -- & -- & -- & -- & 0.80 & 1.20 & 0.40 & -- & -- & -- & -- \\
\midrule
\multirow{4}{*}{Noise} & PGD~\cite{pgd} & 70.00 & 80.00 & 66.40 & 28.31 & 0.668 & 61.1 & 0.194 & 60.85 & 71.10 & 55.50 & 28.33 & 0.649 & 94.7 & 0.316 \\
 & MI-FGSM~\cite{mifgsm} & 75.20 & 82.65 & 69.70 & 27.59 & 0.644 & 80.5 & 0.233 & 68.00 & 76.15 & 62.10 & 27.58 & 0.622 & 109.9 & 0.353 \\
 & TI-DIM~\cite{tidim,dim} & 66.65 & 77.35 & 59.65 & 27.29 & 0.690 & 85.2 & 0.191 & 62.20 & 69.65 & 54.95 & 27.28 & 0.675 & 119.9 & 0.285 \\
 & TIP-IM~\cite{tipim} & 61.85 & 74.75 & 57.75 & \second{32.83} & \second{0.921} & 32.4 & \second{0.053} & 62.60 & 71.55 & 57.85 & \second{32.95} & \second{0.917} & 35.3 & 0.108 \\
\midrule
\multirow{2}{*}{Diffusion} & DiffProtect~\cite{diffprotect} & 34.50 & 49.50 & 32.20 & 24.22 & 0.863 & 33.0 & 0.114 & 35.85 & 46.80 & 28.90 & 25.10 & 0.882 & 49.0 & 0.135 \\
 & WDP~\cite{diffpurify} & \second{90.70} & \second{95.30} & \second{89.70} & 25.38 & 0.777 & \best{22.3} & 0.074 & \second{86.35} & \second{91.45} & \second{83.15} & 26.91 & 0.838 & \best{23.9} & \second{0.080} \\
\midrule
\multirow{3}{*}{Makeup} & AMT-GAN~\cite{amtgan} & 86.60 & 90.70 & 83.75 & 19.19 & 0.720 & 54.9 & 0.151 & 81.10 & 85.90 & 76.50 & 19.31 & 0.699 & 84.9 & 0.221 \\
 & DiffAM~\cite{diffam} & 69.60 & 81.55 & 66.90 & 18.78 & 0.846 & 31.8 & 0.077 & 81.70 & 88.95 & 74.85 & 18.16 & 0.828 & 57.2 & 0.130 \\
 & CLIP2Protect~\cite{clip2protect} & 83.85 & 90.60 & 80.95 & 17.71 & 0.573 & 67.8 & 0.260 & 81.60 & 87.75 & 76.75 & 16.19 & 0.549 & 70.9 & 0.283 \\
\midrule
 & Ours & \best{93.80} & \best{96.70} & \best{92.85} & \best{33.20} & \best{0.957} & \second{28.0} & \best{0.035} & \best{90.65} & \best{94.00} & \best{88.35} & \best{33.29} & \best{0.953} & \second{24.0} & \best{0.058} \\
\bottomrule
\end{tabular}}
\end{table*}

We construct the importance mask once per source image before optimization.
To locate regions where identity similarity is sensitive to random noise,
we add $20$ random probes drawn from $\mathcal U(-\epsilon,\epsilon)$ to the source and,
through the frozen surrogates, compute the gradient of the identity cosine similarity
with respect to each probe.
We average the absolute gradients over probes, sum them across RGB channels,
and apply Gaussian smoothing to obtain a per-pixel importance map,
which concentrates on facial regions.
The top $30\%$ of pixels are assigned $M=1$ and the rest $M=0$.
We regularize neighboring changes in the applied perturbation with masked anisotropic total variation (TV)~\cite{rudin1992tv}.
For horizontal and vertical pairs of adjacent pixels $\mathcal E_h,\mathcal E_v$, let $w_{pq}=M_pM_q$ and define
\begin{equation}
\TV_M(\eta)=
\sum_{d\in\{h,v\}}
\frac{\sum_{c=1}^{3}\sum_{(p,q)\in\mathcal E_d}
w_{pq}|\eta_{c,p}-\eta_{c,q}|}
{\max\!\left\{1,3\sum_{(p,q)\in\mathcal E_d}w_{pq}\right\}} .
\label{eq:tv}
\end{equation}
Each direction averages absolute differences over selected edges and three channels; an empty edge set contributes zero.
Both endpoints must lie inside $M$, so TV regularizes the region's interior rather than its boundary.
Using $\eta$ jointly regularizes the visible effect of both branches.

\subsection{Optimization}
\label{sec:opt}
The main SVD-refined branch carries the principal identity-disrupting perturbation. It uses normalized gradient-direction updates to retain relative entrywise magnitudes, while the auxiliary grid uses sign-PGD for coarse corrections. Both minimize \Cref{eq:loss} using gradients evaluated at the same current branch states. With $g_r=\nabla_{\dhigh}\mathcal L$ and $D=3HW$, define
\begin{equation}
\widehat g_r=
\begin{cases}
\sqrt{D}\,g_r/\|g_r\|_2,&\|g_r\|_2>0,\\
0,&\text{otherwise}.
\end{cases}
\label{eq:grad-direction}
\end{equation}
The norm is computed over all $3HW$ entries of each image.
For $g_r\ne0$, the main update has nominal step length
$\alpha\sqrt{D}$ before refinement.
Both branches use step coefficient $\alpha$: before clipping and refinement, the updates are $\dhigh-\alpha\widehat g_r$ and $\dlow-\alpha\,\mathrm{sign}(\nabla_{\dlow}\mathcal L)$.
The main update applies clipping, channelwise SVD truncation, clipping, Gaussian smoothing, and final clipping, in that order. All three clips use $C_{-\epsilon_r,\epsilon_r}$; the auxiliary grid uses $C_{-\epsilon_g,\epsilon_g}$.
Both branches start from uniform noise within their respective bounds, without refinement. SVDG is applied only once after each main-branch update, without differentiating through SVD. The mask remains fixed, and gradients pass through image formation and its clipping operations. The complete procedure is provided in the appendix; experimental settings are in \Cref{sec:protocol}.

\section{Experiments}
\label{sec:exp}

\subsection{Experimental Setup}
\label{sec:protocol}
\noindent\textbf{Datasets.}
We use two evaluation sets constructed from CelebA-HQ~\cite{celebahq} and VGGFace2~\cite{vggface2}.
Each set contains 500 pairs representing 500 identities, with two different face images of the same identity in each pair.
One image in each pair serves as the source image to be protected, and the other serves as an unprotected reference for face recognition evaluation.
For CelebA-HQ, we use the 500 pairs released with prior work~\cite{diffpurify}; for VGGFace2, we construct 500 pairs through dlib/FFHQ alignment~\cite{kazemi2014alignment,ffhq}.
Protected images are generated at a resolution of $256\times256$, and all compared methods use the same evaluation images and pair assignments.

\begin{table*}[!t]
\centering\footnotesize
\caption{\textbf{Black-box face-swap protection} (PSR, \%). Swapped outputs are verified against the clean source identity; rates average four held-out recognizers. FA denotes Face-Adapter. Bold/underline: best/second best, including ties.}
\label{tab:faceswap}
\setlength{\tabcolsep}{5pt}
\begin{tabular}{llcccc|cccc}
\toprule
& & \multicolumn{4}{c|}{CelebA-HQ} & \multicolumn{4}{c}{VGGFace2} \\
Type & Method & SimSwap & INSwapper & UniFace & FA & SimSwap & INSwapper & UniFace & FA \\
\midrule
& Clean & 0.6 & 0.3 & 1.5 & 2.3 & 1.2 & 1.0 & 6.2 & 4.8 \\
\midrule
\multirow{4}{*}{Noise} & PGD~\cite{pgd} & 65.2 & 50.8 & 79.8 & 79.1 & 56.4 & 46.4 & 87.1 & 76.0 \\
 & MI-FGSM~\cite{mifgsm} & 67.7 & 54.2 & 78.4 & 80.1 & 61.4 & 50.0 & 85.3 & 76.2 \\
 & TI-DIM~\cite{tidim,dim} & 70.6 & 38.9 & 62.5 & 66.0 & 65.4 & 40.3 & 73.1 & 65.5 \\
 & TIP-IM~\cite{tipim} & 66.3 & 42.7 & 70.7 & 65.0 & 65.8 & 48.4 & 81.6 & 70.2 \\
\midrule
\multirow{2}{*}{Diffusion} & DiffProtect~\cite{diffprotect} & 26.0 & 16.9 & 24.9 & 39.6 & 31.9 & 21.3 & 41.6 & 46.4 \\
 & WDP~\cite{diffpurify} & \second{96.5} & \second{82.2} & \second{95.0} & \second{93.8} & \second{94.5} & \second{78.1} & \second{96.4} & \second{93.5} \\
\midrule
\multirow{3}{*}{Makeup} & AMT-GAN~\cite{amtgan} & 90.3 & 76.3 & 90.0 & 91.5 & 86.4 & 72.5 & 92.6 & 89.1 \\
 & DiffAM~\cite{diffam} & 71.8 & 58.3 & 69.0 & 72.1 & 82.9 & 74.4 & 86.9 & 83.1 \\
 & CLIP2Protect~\cite{clip2protect} & 86.0 & 73.9 & 90.5 & 90.5 & 86.5 & 74.9 & 93.9 & 89.9 \\
\midrule
& Ours & \best{98.4} & \best{88.4} & \best{98.5} & \best{94.9} & \best{95.6} & \best{83.5} & \best{99.3} & \best{94.0}    \\
\bottomrule
\end{tabular}
\end{table*}

\noindent\textbf{Target models.}
In each leave-one-out configuration, IR-SE50~\cite{senet}, IR152~\cite{resnet}, FaceNet~\cite{facenet}, and MobileFace~\cite{mobilefacenets} are split into three surrogate recognizers and one held-out evaluator.
All rates average the four held-out results.
For face-swap evaluation, we use four models~\cite{simswap,inswapper,uniface,faceadapter}, identified in \Cref{tab:faceswap}.

\noindent\textbf{Baselines.}
We compare our method with noise-based, diffusion-based, and makeup-based facial privacy protection methods.
Noise-based baselines~\cite{pgd,mifgsm,tidim,dim,tipim} optimize additive adversarial perturbations.
Diffusion-based baselines~\cite{diffprotect,diffpurify} use diffusion models to generate protected portraits.
Makeup-based baselines~\cite{amtgan,clip2protect,diffam} incorporate protection through makeup transfer or makeup-guided editing.
The method-specific generation settings and task adaptations used in the main comparisons are detailed in the appendix. Comparisons under matched $L_\infty$ bounds and iteration counts are reported in \Cref{sec:ablation-studies}, with VGGFace2 results in the appendix.

\noindent\textbf{Implementation details.}
For the main comparisons, we use $256\times256$
input images and a $96\times96$ low-resolution perturbation.
The global $L_\infty$ perturbation budget is
$\epsilon=16/255$, with branch-wise budgets
$\epsilon_r=16/255$ and $\epsilon_g=8/255$.
The SVDG block uses a fixed truncation rank of
$r=11$ and a $3\times3$ Gaussian filter with
$\sigma_r=0.4$.
The importance mask is estimated using 20 probes
and retains the top 30\% of spatial locations;
the masked total-variation weight is $\lambda=20$.
Both branches are jointly optimized for $T=100$
iterations with step size $\alpha=1/255$:
normalized gradient-direction updates for the
full-resolution branch and sign updates for the
low-resolution branch (\Cref{sec:opt}).

\begin{figure*}[t]
\centering
\includegraphics[width=\linewidth]{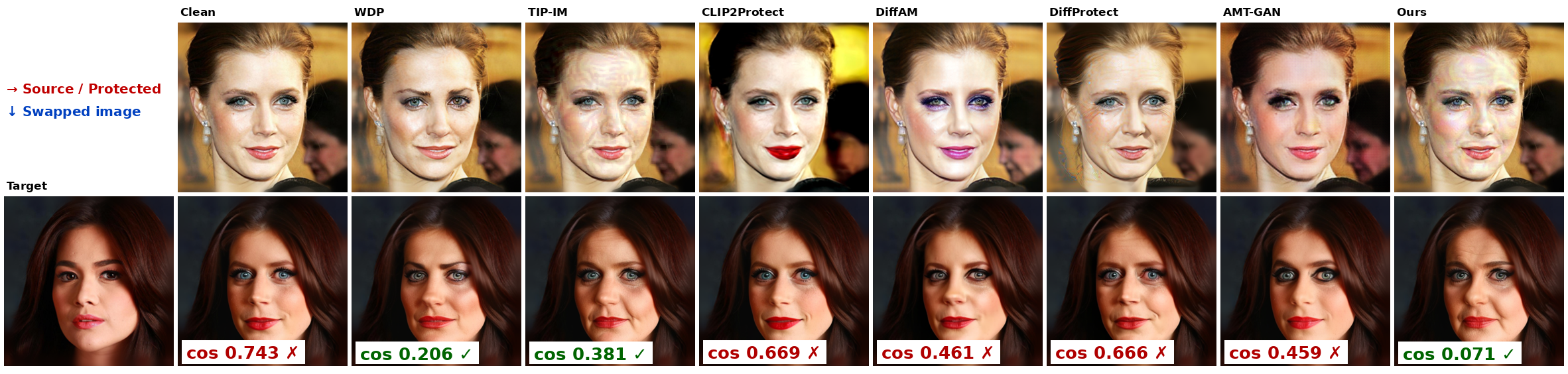}
\caption{\textbf{Qualitative face-swap comparison.} Top: clean or protected sources. Bottom: Face-Adapter outputs using the same target (left), evaluated with held-out FaceNet. Cosine scores compare each output with the clean source; scores below 0.409 indicate successful identity protection.}
\label{fig:qual}
\end{figure*}

\noindent\textbf{Evaluation metrics.}
\textit{Protection performance.}
We report protection success rate (PSR), the fraction of queries rejected as matching the source identity using inherited verification thresholds at nominal FAR $0.01$~\cite{diffpurify}.
Rank-1 and rank-5 protection rates (R1-P/R5-P) measure failure to retrieve the correct reference within the top 1/5 matches in a 500-image gallery.
Higher values indicate stronger protection; detection failures count as protection failures.
Detailed evaluation protocols are provided in the appendix.

\noindent\textit{Visual quality.}
We assess protected-image quality using PSNR, SSIM~\cite{ssim}, FID~\cite{fid}, and LPIPS~\cite{lpips}.
PSNR, SSIM, and LPIPS compare each protected image with its clean source, while FID compares the two image collections.
Higher PSNR/SSIM and lower FID/LPIPS indicate better visual quality.

\subsection{Comparison Study}
\label{sec:comparison}
\paragraph{Black-Box Dodging}
\label{sec:dodging}
Face-swapping methods use face-recognition embeddings to transfer source identity~\cite{simswap,faceadapter}.
We therefore evaluate black-box dodging on held-out recognizers to assess identity disruption relevant to face-swap protection.
In \Cref{tab:dodging}, DUNE has the highest dodging and retrieval protection, PSNR, and SSIM, and the lowest LPIPS on both datasets.
WDP has lower FID, so DUNE's quality advantage is clearest in source-specific preservation.

\noindent\textbf{Black-Box Face-Swap Protection}
\label{sec:faceswap}
For each dataset, a random permutation assigns each source a clean target of a different identity; these assignments are fixed across all protection methods and the four black-box swappers in \Cref{sec:protocol}.
Each swapped output is compared with its original clean source using the held-out recognizer.
Protection is successful when their identity embeddings have cosine similarity below that recognizer's verification threshold.
DUNE achieves the highest PSR among compared methods in all eight dataset--swapper combinations (\Cref{tab:faceswap}), with the best source PSNR, SSIM, and LPIPS (\Cref{tab:dodging}).
\Cref{fig:qual} illustrates protected sources and their swapped outputs; examples are provided in the appendix.

\subsection{Ablation and Controlled Comparisons}
\label{sec:ablation-studies}
The component ablations use CelebA-HQ under the settings in \Cref{sec:protocol}, with the indicated components added or removed. SVDG denotes the block in \Cref{fig:pipeline}(b); importance masking and masked TV are separate components. We also compare methods under matched perturbation and iteration budgets.

\noindent\textbf{Comparison with Frequency Filtering}
\Cref{tab:refinement} compares SVDG with fixed-frequency filtering, including the DCT low-pass filtering strategy used in FaceShield~\cite{jeong2025faceshield}, and with each refinement step in isolation. SVD truncation at rank 11 improves all four source-fidelity metrics over DCT and FFT filtering. Combining truncation with Gaussian smoothing achieves the best fidelity among the tested variants. Compared with Gaussian smoothing alone, SVDG reduces FID from 44.4 to 30.2 and LPIPS from 0.0756 to 0.0587, with dodging PSR decreasing from 93.40\% to 90.55\%.

\begin{table}[!t]
\centering\footnotesize
\caption{\textbf{Frequency filtering and SVDG on CelebA-HQ.} $\epsilon=16/255$, 100 normalized-gradient steps, and $\alpha=1/255$; no masking, TV, or auxiliary branch. Dodging PSR (\%) averages four held-out recognizers. FID parentheses show improvement ($-$) or degradation ($+$) relative to None.}
\label{tab:refinement}
\setlength{\tabcolsep}{2pt}
\begin{tabular*}{\columnwidth}{@{\extracolsep{\fill}}lccccc@{}}
\toprule
Condition & PSR$\uparrow$ & PSNR$\uparrow$ & SSIM$\uparrow$ & LPIPS$\downarrow$ & FID$\downarrow$ \\
\midrule
None & 86.10 & 27.78 & 0.652 & 0.2160 & 70.2\,\textcolor{black}{\scriptsize$(0.0)$} \\
DCT Q40 ~\cite{jeong2025faceshield} & 91.50 & 29.65 & 0.765 & 0.1804 & 71.4\,\metricgain{1.2} \\
FFT ($\rho\leq0.4$) & 92.60 & 30.20 & 0.800 & 0.1586 & 71.6\,\metricgain{1.4} \\
SVD rank 11 & 85.50 & 31.47 & 0.854 & 0.0975 & 45.3\,\metricloss{24.9} \\
Gaussian ($\sigma=0.4$) & \best{93.40} & 32.08 & 0.908 & 0.0756 & 44.4\,\metricloss{25.8} \\
SVDG (Ours) & 90.55 & \best{32.77} & \best{0.926} & \best{0.0587} & \best{30.2}\,\metricloss{40.0} \\
\bottomrule
\end{tabular*}
\end{table}

\begin{table}[!t]
\centering\footnotesize
\caption{\textbf{Cumulative ablation.} Components are added to the SVDG full-resolution base in row order. PGD uses 100 normalized-gradient steps without refinement, masking, or TV. Dodging PSR (\%) averages four held-out recognizers.}
\label{tab:cumulative}
\setlength{\tabcolsep}{4pt}
\begin{tabular}{lcccc}
\toprule
Configuration \textbackslash{} Dodging & PSR$\uparrow$ & PSNR$\uparrow$ & FID$\downarrow$ & LPIPS$\downarrow$ \\
\midrule
PGD (reference) & 86.10 & 27.78 & 70.2 & 0.216 \\
\midrule
SVDG (base) & 90.55 & 32.77 & 30.2 & 0.059 \\
+ Importance mask & 88.90 & 34.10 & 24.4 & 0.036 \\
+ Masked TV & 86.40 & \best{35.08} & \best{17.8} & \best{0.023} \\
+ Auxiliary branch (Ours) & \best{93.80} & 33.20 & 28.0 & 0.035 \\
\bottomrule
\end{tabular}
\end{table}

\begin{table}[!t]
\centering\footnotesize
\caption{\textbf{Component ablation.} Each ablation removes only the named component from Full. w/o SVD keeps the Gaussian smoothing of SVDG. Face-swap PSR (\%) averages four swappers and four held-out recognizers.}
\label{tab:component}
\setlength{\tabcolsep}{3.5pt}
\begin{tabular}{@{}lcccc@{}}
\toprule
Configuration \textbackslash{} Face-swap & PSR$\uparrow$ & PSNR$\uparrow$ & SSIM$\uparrow$ &FID$\downarrow$ \\
\midrule
Full (Ours) & 95.1 & 33.20 & 0.957 & 28.0 \\
\midrule
w/o Auxiliary branch & 88.0 & 35.08 & 0.968 & 17.8 \\
w/o SVD & 96.2 & 32.22 & 0.946 & 40.8 \\
w/o SVDG & 96.3 & 30.34 & 0.919 & 81.0 \\
w/o Importance mask (\cref{fig:pipeline}(d)) & 95.7 & 30.64 & 0.863 & 52.9 \\
w/o Masked TV loss & 95.3 & 32.76 & 0.932 & 36.4 \\
\bottomrule
\end{tabular}
\end{table}

\begin{figure*}[t]
\centering
\includegraphics[width=\linewidth]{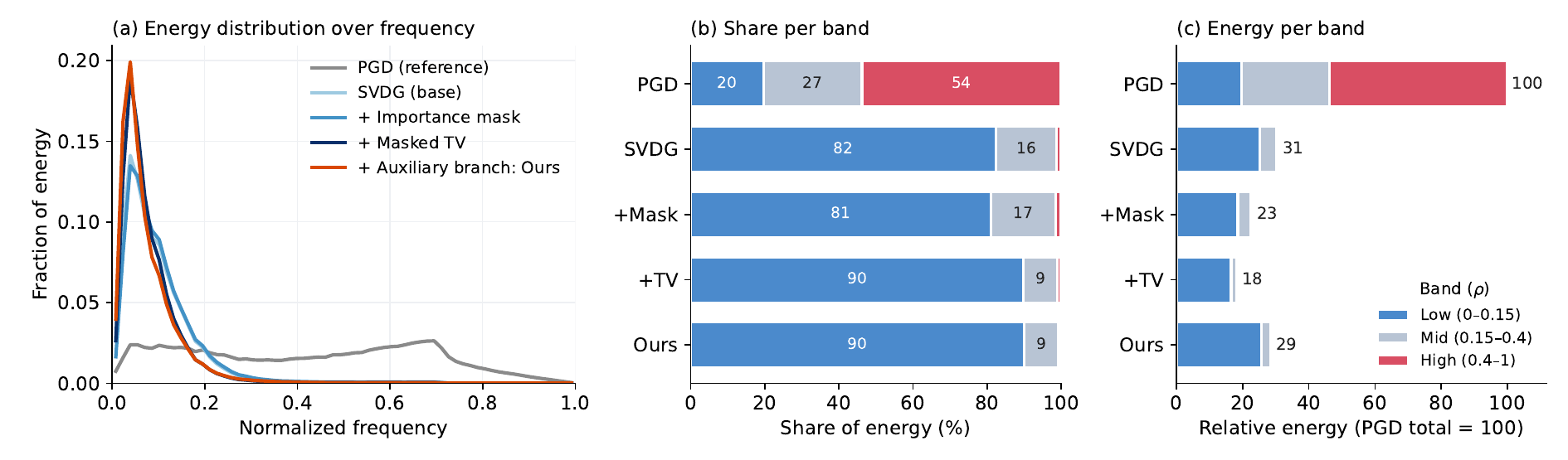}
\caption{\textbf{Frequency distribution of the perturbation energy for the configurations in \Cref{tab:cumulative}.} (a)~Fraction of the energy of $\eta=x_{\mathrm p}-x$ in each of 64 equal-width frequency bins (values sum to 1). (b)~Share of the energy in the low ($[0,0.15)$), mid ($[0.15,0.4)$), and high ($[0.4,1]$) bands. (c)~Energy per band relative to the total energy of PGD (100).}
\label{fig:cum-spectrum}
\end{figure*}

\noindent\textbf{Component Contributions}
\label{sec:ablation}
\Cref{tab:cumulative,tab:component} show how the components contribute to protection and source preservation. Compared with the SVDG base, the complete method increases dodging PSR from 90.55\% to 93.80\%, improves PSNR from 32.77 to 33.20 dB, and reduces LPIPS from 0.059 to 0.035 (\Cref{tab:cumulative}). Importance masking and masked TV reduce distortion, while the auxiliary branch recovers protection under these spatial constraints. In \Cref{tab:component}, adding the auxiliary branch increases mean face-swap PSR from 88.0\% to 95.1\%, while source PSNR decreases from 35.08 to 33.20 dB. Removing SVDG from the full method increases mean PSR to 96.3\%, while reducing source PSNR from 33.20 to 30.34 dB and increasing FID from 28.0 to 81.0. Removing only importance masking or only the masked TV loss also worsens source fidelity: FID rises from 28.0 to 52.9 or 36.4, respectively, with little change in mean PSR. \Cref{fig:cum-spectrum} shows the corresponding frequency profiles using the bands in \Cref{fig:component}. SVDG reduces the high-band energy share, while masking and masked TV further reduce total energy. The configuration with the auxiliary branch has more low-band energy alongside the recovery of protection, consistent with its role in providing spatially coupled corrections.

\noindent\textbf{Comparison under Matched Budgets}
\label{sec:budget-tradeoff}
We compare DUNE with noise-based baselines~\cite{pgd,mifgsm,tidim,dim,tipim} under the budget conventionally used for PGD, MI-FGSM, and TI-DIM.
Update directions remain method-specific, as detailed in the appendix.
\begin{table}[t]
\centering
\caption{\textbf{Face-swap protection under matched bounds and iteration counts.} CelebA-HQ; $\epsilon=16/255$, 10 iterations, and $\alpha=1.6/255$. PSR (\%) averages four held-out recognizers; update directions remain method-specific.}
\label{tab:budget-faceswap}
\footnotesize
\setlength{\tabcolsep}{2pt}
\begin{tabular*}{\columnwidth}{@{\extracolsep{\fill}}l*{8}{c}@{}}
\toprule
Method & SimSwap & INSw. & UniFace & FA & Avg. & PSNR$\uparrow$ & SSIM$\uparrow$ & FID$\downarrow$ \\
\midrule
Clean & 0.6 & 0.3 & 1.5 & 2.3 & 1.2 & -- & -- & -- \\
PGD & 65.2 & 50.8 & 79.8 & 79.1 & 68.7 & 28.31 & 0.668 & 61.1 \\
MI-FGSM & 67.7 & 54.2 & 78.4 & \second{80.1} & 70.1 & 27.59 & 0.644 & 80.5 \\
TI-DIM & 70.6 & 38.9 & 62.5 & 66.0 & 59.5 & 27.29 & 0.690 & 85.2 \\
TIP-IM & \second{82.3} & \second{61.0} & \second{83.0} & \second{80.1} & \second{76.6} & \second{30.06} & \second{0.846} & \second{60.3} \\
\midrule
Ours & \best{93.6} & \best{81.5} & \best{93.5} & \best{90.2} & \best{89.7} & \best{35.44} & \best{0.964} & \best{19.6} \\
\bottomrule
\end{tabular*}
\end{table}
\Cref{tab:budget-faceswap} shows stronger face-swap protection and better
source fidelity for DUNE under the matched settings. DUNE leads on all four
swappers; compared with TIP-IM, the average face-swap PSR rises from 76.6\%
to 89.7\% while FID falls from 60.3 to 19.6. Dodging results under the same
settings show the same ordering (88.70\% versus 79.40\% PSR; see the appendix).
The joint improvement indicates that the protection benefit persists
when the perturbation bound and iteration count are matched.
VGGFace2 results and visual comparisons are provided in the appendix.

\section{Limitations}
\label{sec:limitations}
We evaluate on two aligned portrait datasets, four face-swapping systems, and four held-out recognizers at $256\times256$ resolution.
Our perturbation is optimized only against face recognizers.
It may transfer less well to face-swapping models that extract identity differently, which we did not test.
The observed frequency trends and idealized analysis do not guarantee high-frequency suppression for arbitrary PGD perturbations.
Refinement can trade protection for fidelity, and the fixed GD-calibrated rank is not established as optimal.

\section{Conclusion}
\label{sec:conclusion}
We presented DUNE to protect portraits against face swapping while preserving their appearance. Our frequency analysis and controlled rank experiments motivate selective SVD truncation during optimization.
DUNE jointly optimizes a main full-resolution branch, refined by SVDG (truncation followed by Gaussian smoothing), and an auxiliary low-resolution branch that complements protection. Importance masking and masked TV then regulate the spatial placement and local variation of their combined perturbation.
Experiments on two datasets and four black-box swappers demonstrate strong protection and source fidelity, with ablations supporting the complementary roles of refinement, spatial regularization, and the auxiliary branch.

\clearpage
\section*{LLM Usage Statement}
OpenAI Codex and Anthropic Claude were used to assist in editing and refining author-produced manuscript text, code, and data-related materials. AI tools were also used to generate figures and diagrams, which the authors reviewed, edited, or approved. Reviewer-style feedback on the manuscript or substantial parts of it was obtained using OpenAI models. The authors take full responsibility for all submitted content, regardless of any LLM or AI-tool use.

{
    \small
    \bibliographystyle{ieeenat_fullname}
    \bibliography{main}
}

\clearpage
\appendix
\raggedbottom
\clearpage
\setcounter{equation}{0}
\renewcommand{\theequation}{S\arabic{equation}}
\setcounter{theorem}{0}
\renewcommand{\thetheorem}{S\arabic{theorem}}
\setcounter{section}{0}
\renewcommand{\thesection}{\Alph{section}}
\renewcommand{\theHsection}{supp.\arabic{section}}
\setcounter{table}{0}
\renewcommand{\thetable}{S\arabic{table}}
\renewcommand{\theHtable}{supp.\arabic{table}}
\setcounter{figure}{0}
\renewcommand{\thefigure}{S\arabic{figure}}
\renewcommand{\theHfigure}{supp.\arabic{figure}}
\setcounter{algorithm}{0}
\renewcommand{\thealgorithm}{S\arabic{algorithm}}
\providecommand{\theHalgorithm}{supp.\arabic{algorithm}}
\maketitlesupplementary

This appendix provides theoretical analyses, evaluation details, and additional experiments for DUNE.
\Cref{sec:suppl_theory} covers GD-based rank calibration, sufficient conditions for high-frequency removal by SVD truncation, and the rank, frequency, and representation properties of the two branches, including proofs of the main-paper propositions.
\Cref{sec:supp-null,sec:supp-component-256} report empirical rank estimates and singular-component frequency profiles at the evaluation resolution.
\Cref{sec:supp-protocol} describes evaluation and generation settings, comparisons with face-swap defenses, and additional qualitative examples.
\Cref{sec:isobudget,sec:supp-iterations} examine protection and source fidelity under matched budgets and varying iteration counts.
\Cref{sec:supp-joint} examines the role of the low-resolution branch by comparing the standalone branches with the complete method.
\Cref{sec:supp-svd-gauss,sec:supp-svd-timing} illustrate how Gaussian smoothing reduces SVD truncation artifacts and compare post-hoc with per-step truncation.
\Cref{sec:supp-svd-freeze} analyzes how SVD subspaces vary across images and evolve during optimization, and evaluates the effect of freezing the truncation bases.
Finally, \Cref{sec:transform-robustness} evaluates robustness to image transformations and diffusion purification, \Cref{sec:supp-cost} reports generation cost, and \Cref{sec:supp-algorithm} presents the complete optimization procedure.

\section{Theoretical Analysis of DUNE}
\label{sec:suppl_theory}

\subsection{Scope and Notation}
\label{sec:suppl_scope_and_notation}

We analyze square image channels.
Let $N$ and $M$, with $M\ge N$, denote the
low-resolution and full-resolution side lengths.
For a low-resolution channel
$\delta_g^{(c)}\in\mathbb{R}^{N\times N}$,
bilinear upsampling takes the form
\begin{equation}
  U(\delta_g^{(c)})
  =
  A\delta_g^{(c)}A^\top,
  \qquad
  A\in\mathbb{R}^{M\times N}.
  \label{eq:dune-upsample}
\end{equation}
The interpolation weights are defined in~\eqref{eq:dune-A}.

Following the main paper, let
$\delta_r^{\mathrm{pre}}\in\mathbb{R}^{M\times M}$
denote a full-resolution channel immediately before
SVD truncation, with its channel index omitted.
We write $\mathcal P_r(X)$ for the reconstruction
from the leading $r$ singular components of $X$.
Singular values $s_i(X)$ are ordered nonincreasingly.
We use $\|X\|_2$ for the operator norm and
$\|X\|_F$ for the Frobenius norm.
Pixel and Fourier indices start at zero,
whereas singular-value indices start at one.
The effective perturbation after complete image
formation is $\eta=x_p-x$.
The analysis concerns the specified interpolation
and truncation stages; its scope under subsequent
processing is discussed in the final remark.

We first describe the GD-based rank calibration
and establish a sufficient condition under which
SVD truncation removes high-frequency components.
We then characterize the rank and frequency
constraints of bilinear upsampling and show how
combining the two branches expands the class
of representable perturbations.

\subsection{GD-Based Rank Selection}
\label{sec:suppl_rmt_threshold}

\paragraph{Reference threshold.}
For a square matrix $Y\in\mathbb{R}^{M\times M}$,
let $s_{\mathrm{med}}(Y)$ denote the median of its
singular values.
The Gavish--Donoho (GD) hard threshold for an
unknown noise level is
\begin{equation}
  \widehat{\tau}_{\mathrm{GD}}(Y)
  \approx
  2.858\,s_{\mathrm{med}}(Y).
  \label{eq:dune-gd-unknown}
\end{equation}
The coefficient is derived from asymptotically
optimal hard thresholding for squared Frobenius
reconstruction loss under the low-rank-plus-white-noise
assumptions of Gavish and Donoho~\citep{gavish2014optimal}.

We use this threshold to estimate the number of
components to retain:
\begin{equation}
  \widehat r(Y)
  =
  \#\left\{
    i:s_i(Y)>\widehat{\tau}_{\mathrm{GD}}(Y)
  \right\}.
  \label{eq:dune-rmt-rank}
\end{equation}

\paragraph{Calibration on PGD perturbations.}
We apply~\eqref{eq:dune-rmt-rank} separately to each
$256\times256$ color channel of the effective PGD
perturbation $\eta=x_p-x$. Calibration uses 500 CelebA-HQ
identities disjoint from the evaluation set. Across four
held-out recognizer configurations and three color channels,
the resulting 6{,}000 rank estimates have a pooled median
of 11. We therefore fix $r=11$ throughout optimization.
\Cref{tab:supp-rank-seeds} reports the estimates for each
of the five calibration samples.

We use the calibrated rank in all DUNE experiments.
The next subsection gives a sufficient condition for
truncation at a given rank to recover the signal component.

\subsection{SVD Recovery Under Orthogonal Frequency Support}
\label{sec:suppl_svd_hf}

We prove \Cref{prop:main-svd-hf} of the main paper,
which establishes a sufficient condition for SVD
truncation to remove an HF component while
preserving a low-rank LF component.

Let $P_H\in\mathbb{R}^{M\times M}$ be a fixed
orthogonal projector onto an $m$-dimensional
HF subspace, where $1\le m<M$.
Consider
\begin{equation}
  \begin{aligned}
    \delta_r^{\mathrm{pre}}
    &=S+E,
    \qquad E=P_HGP_H,\\
    S&=(I_M-P_H)S(I_M-P_H),
  \end{aligned}
  \label{eq:suppl-svd-model}
\end{equation}
where $\operatorname{rank}(S)=r$ and
$1\le r\le M-m$.
The Gaussian reference model in the main paper
takes the entries of $G$ independently from
$\mathcal N(0,\sigma^2/m)$, with $\sigma>0$.

\begin{proof}[Proof of main-paper \Cref{prop:main-svd-hf}]
Choose fixed orthonormal bases $V_H$ and $V_L$ for the range of $P_H$ and its orthogonal complement, respectively.
Then $V=[V_L,V_H]$ is orthogonal, and
\begin{equation}
  V^\top\delta_r^{\mathrm{pre}}V
  =
  \begin{pmatrix}
    S_L & 0\\
    0 & E_H
  \end{pmatrix},
  \label{eq:suppl-svd-block}
\end{equation}
where
$S_L=V_L^\top SV_L$ and
$E_H=V_H^\top GV_H$.

Orthogonal multiplication preserves singular values.
The singular values of a block-diagonal matrix
are the combined singular values of its
diagonal blocks.
The $r$ nonzero singular values of $S_L$ are those
of $S$, and every singular value of $E_H$ is at most
$\|E_H\|_2=\|E\|_2$.

Consequently, if $s_r(S)>\|E\|_2$, the leading
$r$ singular components all belong to the $S_L$
block.
Truncating to these components and transforming
back therefore gives
\[
  \mathcal P_r(\delta_r^{\mathrm{pre}})=S.
\]
\end{proof}

\paragraph{Scale of the Gaussian HF component.}
The recovery argument relies on orthogonal support
and a singular-value gap, while the Gaussian
assumption provides a reference scale for
$\|E\|_2$.
For this fixed basis, orthogonal invariance of the Gaussian
matrix $G$ gives
\begin{equation}
  E_H
  \overset{\mathrm d}{=}
  \frac{\sigma}{\sqrt m}\,\Xi_m,
  \label{eq:suppl-gaussian-hf}
\end{equation}
where $\Xi_m$ is an $m\times m$ matrix with
independent standard Gaussian entries.
The square-Gaussian noise-edge limit therefore gives
\begin{equation}
  \|E\|_2
  \longrightarrow 2\sigma
  \qquad\text{almost surely as }m\to\infty
  \label{eq:suppl-gaussian-edge}
\end{equation}
along a sequence of such models~\citep{gavish2014optimal}.

This limit provides a quantitative interpretation
of the recovery condition in \Cref{prop:main-svd-hf}.
In particular, if the weakest retained signal
component satisfies $s_r(S)\ge 2\sigma+\gamma$
for some fixed margin $\gamma>0$ along the
sequence, then
\begin{equation}
  \mathbb P\!\left(s_r(S)>\|E\|_2\right)
  \longrightarrow 1
  \qquad\text{as }m\to\infty.
  \label{eq:suppl-recovery-probability}
\end{equation}
Under the Gaussian scaling in~\eqref{eq:suppl-gaussian-hf},
the signal strength required for separation need not grow
with the number of HF modes. Even a fixed value of $s_r(S)$
above $2\sigma$ suffices for exact recovery with probability
tending to one under the assumed orthogonal support.
For a finite channel, \Cref{prop:main-svd-hf} compares $s_r(S)$ with
the realized largest HF singular value $\|E\|_2$.

\paragraph{Relation to the frequency definition.}
For even $M$, a concrete choice consistent with
the radial FFT convention in
\Cref{sec:suppl_frequency_bilinear}
is to let $P_H$ retain the one-dimensional Fourier
modes with $|\kappa(u)|>M/5$.
With this choice, $S$ is supported in $\rho\le0.4$, which
includes the low- and mid-frequency bands of \Cref{fig:component},
and $E$ is supported entirely in $\rho>0.4$.
To obtain a signal confined to the low-frequency band of
\Cref{fig:component}, one can further restrict the Fourier support of
$S$ to $\rho<0.15$. The same recovery proof applies.
The model excludes mixed low--high and high--low blocks.

\subsection{Rank Preservation Under Bilinear Upsampling}
\label{sec:suppl_rank_bilinear}

We specify the interpolation operator and prove
the rank-preservation statement used in
\Cref{sec:lowres} of the main paper.

\paragraph{Interpolation operator.}
\label{sec:suppl_interpolation}
We use half-pixel coordinates with edge-value
extension, corresponding to bilinear resizing with
\texttt{align\_corners=False} and
\texttt{antialias=False} when the output size is
explicitly prescribed~\citep{onnxresize,torchinterpolate}.
For $i=0,\ldots,M-1$, define
\begin{equation}
  \begin{aligned}
    t_i
    &=\frac{N}{M}\left(i+\frac12\right)-\frac12,\\
    q_i
    &=\min\{N-1,\max\{0,t_i\}\},\\
    A_{ij}
    &=(1-|q_i-j|)_+,
    \qquad j=0,\ldots,N-1,
  \end{aligned}
  \label{eq:dune-A}
\end{equation}
where $(a)_+=\max\{a,0\}$.
Each row has at most two nonzero entries,
$A_{ij}\ge0$, and
$A\mathbf 1_N=\mathbf 1_M$.
Applying $A$ along both spatial axes gives
\eqref{eq:dune-upsample}.

\begin{lemma}[Full Column Rank of Half-Pixel Upsampling]
\label{lem:dune-fullrank}
For integers $M\ge N\ge1$, the matrix
in~\eqref{eq:dune-A} has full column rank $N$.
\end{lemma}

\begin{proof}
If $M=N$, then $A=I_N$.
Suppose $M>N$.
For each input index $j$, select an output index
$i_j$ nearest to
\[
  \xi_j
  =
  \frac{M}{N}\left(j+\frac12\right)-\frac12.
\]
Since $\xi_j\in[0,M-1]$, we have
$|i_j-\xi_j|\le1/2$, and hence
\[
  |t_{i_j}-j|
  \le\frac{N}{2M}
  <\frac12.
\]
Clamping to $[0,N-1]$ cannot increase the
distance to $j$, so $A_{i_jj}>1/2$.
The selected indices are distinct because
a single clamped coordinate cannot be within
distance strictly less than $1/2$ of two
distinct integers.

The square submatrix formed by these rows,
ordered by $j$, is strictly row diagonally
dominant: each diagonal entry exceeds $1/2$,
whereas the remaining entries in that row
sum to less than $1/2$.
Such a matrix is invertible.
Indeed, a nonzero null vector would violate
the row equation corresponding to one of its
largest absolute coordinates.
Thus $A$ contains an invertible
$N\times N$ submatrix.
\end{proof}

\paragraph{Rank preservation.}
\begin{proposition}[Rank Preservation]
\label{prop:dune-rank}
Let $A\in\mathbb{R}^{M\times N}$ have full
column rank.
For every channel
$\delta_g^{(c)}\in\mathbb{R}^{N\times N}$,
\begin{equation}
  \operatorname{rank}\!\left(U(\delta_g^{(c)})\right)
  =
  \operatorname{rank}(\delta_g^{(c)})
  \le N.
  \label{eq:dune-rank}
\end{equation}
\end{proposition}

\begin{proof}
Since $A$ has full column rank, its pseudoinverse
satisfies $A^\dagger A=I_N$.
Therefore,
\[
  \delta_g^{(c)}
  =
  A^\dagger U(\delta_g^{(c)})(A^\dagger)^\top.
\]
Matrix multiplication cannot increase rank.
Applying this fact to the identity above and
to~\eqref{eq:dune-upsample} proves equality
of the input and output ranks.
\end{proof}

For $N=96$ and $M=256$, the output rank is
at most 96, so at least 160 output singular
values are zero.
Bilinear upsampling therefore increases spatial
resolution without increasing matrix rank.

For the frequency and representation analyses
below, define
\begin{equation}
  Q=A(A^\top A)^{-1/2},
  \qquad
  Q^\top Q=I_N.
  \label{eq:dune-polar}
\end{equation}
Thus $Q$ is an orthonormal basis for
$\operatorname{col}(A)$.

\begin{remark}[Rectangular Grids]
For a channel
$Z\in\mathbb{R}^{N_h\times N_w}$ and
full-column-rank interpolation matrices
$A_h$ and $A_w$, the same argument gives
\[
  \operatorname{rank}(A_hZA_w^\top)
  =
  \operatorname{rank}(Z).
\]
\end{remark}

\subsection{High-Frequency Energy Under Bilinear Upsampling}
\label{sec:suppl_frequency_bilinear}

We derive the worst-case HF energy fraction
reported in \Cref{sec:lowres} of the main paper.
The bound holds for every nonzero real-valued input grid
and requires no distributional assumption.

\paragraph{FFT convention and HF energy.}
Let $M$ be even and let
$F_M\in\mathbb{C}^{M\times M}$ be the unitary
discrete Fourier transform matrix,
\begin{equation}
  (F_M)_{uj}
  =
  \frac{1}{\sqrt M}
  \exp\!\left(-\frac{2\pi\mathrm{i}uj}{M}\right),
  \label{eq:suppl-dft}
\end{equation}
where $u,j=0,\ldots,M-1$.
For a real channel $X$, its two-dimensional
Fourier transform is
\[
  \widehat X=F_MXF_M^\top.
\]
The transpose on the right applies the forward
transform along the second spatial axis.

Define the signed frequency index
\begin{equation}
  \kappa(u)
  =
  \begin{cases}
    u,   & 0\le u<M/2,\\
    u-M, & M/2\le u<M.
  \end{cases}
  \label{eq:suppl-fft-index}
\end{equation}
Following the main-paper frequency analysis,
the normalized radial frequency and HF mask are
\begin{equation}
  \begin{aligned}
    \rho_{uv}
    &=
    \frac{\sqrt{\kappa(u)^2+\kappa(v)^2}}
         {M/\sqrt{2}},\\
    H_{uv}
    &=
    \mathbf 1_{\{\rho_{uv}>0.4\}}.
  \end{aligned}
  \label{eq:suppl-fft-mask}
\end{equation}
For $X\ne0$, define its HF energy fraction as
\begin{equation}
  \operatorname{HF}(X)
  =
  \frac{
    \|H\odot\widehat X\|_F^2
  }{
    \|X\|_F^2
  }.
  \label{eq:suppl-hf-fraction}
\end{equation}
The denominator equals
$\|\widehat X\|_F^2$ by unitarity.

\paragraph{Sharp bound on the interpolation space.}
Let $\operatorname{vec}$ stack matrix columns,
let $\otimes$ denote the Kronecker product,
and let $^*$ denote conjugate transpose.
Using $Q$ from~\eqref{eq:dune-polar}, define
\begin{equation}
  \begin{aligned}
    V&=Q\otimes Q,
    &F_2&=F_M\otimes F_M,\\
    D_H&=\operatorname{diag}(\operatorname{vec}(H)),
    &K&=V^\top F_2^*D_HF_2V.
  \end{aligned}
  \label{eq:suppl-hf-operator}
\end{equation}

\begin{proposition}[Sharp HF Bound for Bilinear Upsampling]
\label{prop:suppl-fft-hf}
Let $A\in\mathbb{R}^{M\times N}$ have full
column rank, and use the frequency mask
in~\eqref{eq:suppl-fft-mask}.
Then $K$ is real symmetric,
$0\preceq K\preceq I_{N^2}$, and
\begin{equation}
  \sup_{\substack{
    Z\in\mathbb{R}^{N\times N}\\ Z\ne0
  }}
  \operatorname{HF}(AZA^\top)
  =
  c(A)
  :=
  \lambda_{\max}(K).
  \label{eq:suppl-sharp-hf}
\end{equation}
In particular, every nonzero grid channel satisfies
\[
  \operatorname{HF}\!\left(U(\delta_g^{(c)})\right)
  \le c(A).
\]
\end{proposition}

\begin{proof}
For an arbitrary grid $Z$, set
\[
  W=(A^\top A)^{1/2}Z(A^\top A)^{1/2}.
\]
Then
\[
  AZA^\top=QWQ^\top,
  \qquad
  \|QWQ^\top\|_F=\|W\|_F.
\]
Writing $w=\operatorname{vec}(W)$ gives
\[
  \operatorname{vec}(QWQ^\top)=Vw,
  \qquad
  \operatorname{vec}(\widehat{QWQ^\top})=F_2Vw.
\]

Since $D_H$ is an orthogonal coordinate selector
and $V$ has orthonormal columns,
$0\preceq K\preceq I_{N^2}$.
The radial mask satisfies
\[
  H_{uv}
  =
  H_{(-u)\bmod M,\,(-v)\bmod M}.
\]
This conjugate symmetry makes
$F_2^*D_HF_2$ real symmetric.
Consequently, $K$ is also real symmetric.

The HF fraction is therefore the Rayleigh quotient
\begin{equation}
  \operatorname{HF}(AZA^\top)
  =
  \frac{w^\top Kw}{w^\top w}
  \le\lambda_{\max}(K).
  \label{eq:suppl-hf-rayleigh}
\end{equation}
The change of variables between $Z$ and $W$
is invertible.
Choose a real unit eigenvector $w$ corresponding
to $\lambda_{\max}(K)$, reshape it into $W$,
and set
\[
  Z=(A^\top A)^{-1/2}W(A^\top A)^{-1/2}.
\]
This real, nonzero grid attains equality,
which proves sharpness.
\end{proof}

\paragraph{Evaluation for $96\times96$ to $256\times256$.}
For the interpolation matrix in~\eqref{eq:dune-A}
with $N=96$ and $M=256$, numerical evaluation
in float64 gives
\begin{equation}
  c(A)\approx0.3237931569.
  \label{eq:suppl-hf-numeric}
\end{equation}
For this interpolation operator, at most $32.38\%$ of the
output energy lies in the HF band. The full-resolution rank-one
channel $X_{ij}=(-1)^i$ has all its Fourier energy at the
Nyquist frequency $(M/2,0)$ and therefore lies outside the
interpolation space.

The largest eigenvalue can be evaluated without
constructing the full matrix $K$.
For a vector reshaped into $W$, first form
$QWQ^\top$, apply the two-dimensional FFT,
retain the coefficients selected by $H$,
and apply the inverse FFT.
Multiplying the result by $Q^\top$ and $Q$
and vectorizing gives $Kw$.
A symmetric iterative eigensolver then evaluates
$\lambda_{\max}(K)$.
An orthonormal basis obtained from a thin QR
factorization of $A$ can replace $Q$ without
changing this eigenvalue.

Since the HF energy fraction is invariant under nonzero
rescaling, a real grid attaining the bound can be scaled
to satisfy any positive entrywise grid budget $\epsilon_g$.
The bound applies to the interpolated channel before branch
addition, masking, and clipping of the combined perturbation.

\subsection{Complementary Representation Spaces}
\label{sec:suppl_dual_representation}

The two branches of DUNE impose distinct structural
constraints. SVD truncation retains perturbation-dependent
singular directions but restricts the matrix rank to $r$.
Bilinear upsampling restricts the perturbation to a fixed
interpolation space but permits ranks up to $N$, while
limiting its HF energy fraction
(\Cref{prop:suppl-fft-hf}).
Their combination therefore provides additional
representational flexibility without increasing the
truncation rank of the main branch.
We formalize this complementarity in
\Cref{prop:main-dual-expressivity} of the main paper.

The SVD-truncated channel belongs to
\begin{equation}
  \mathcal R_r
  =
  \left\{
    X\in\mathbb{R}^{M\times M}:
    \operatorname{rank}(X)\le r
  \right\}.
  \label{eq:dune-Rr}
\end{equation}
The bilinearly interpolated channel belongs to
the fixed linear space
\begin{equation}
  \mathcal I_A
  =
  \left\{
    AZA^\top:
    Z\in\mathbb{R}^{N\times N}
  \right\}.
  \label{eq:dune-IA}
\end{equation}
Every matrix in $\mathcal I_A$ has its row and
column spaces contained in $\operatorname{col}(A)$.
Their sum $\mathcal R_r+\mathcal I_A$ represents
the combined class at these two stages.

\begin{proposition}[Complementary Representation Spaces]
\label{prop:dune-dual}
Suppose $1\le r<N<M$ and $A$ has full column rank.
Then neither $\mathcal R_r$ nor $\mathcal I_A$
contains the other, and
\begin{equation}
  \mathcal R_r\cup\mathcal I_A
  \subsetneq
  \mathcal R_r+\mathcal I_A.
  \label{eq:suppl-dual-space-sum}
\end{equation}
Moreover, for
$X\in\mathcal I_A$ and $Y\in\mathcal R_r$,
\begin{equation}
  \operatorname{rank}(X+Y)
  \le
  \min\{M,N+r\}.
  \label{eq:dune-dual-rank}
\end{equation}
\end{proposition}

\begin{proof}
Let $X=AA^\top$ and choose a unit vector
$v\perp\operatorname{col}(A)$.
Then $X\in\mathcal I_A$ has rank $N>r$,
whereas $Y=vv^\top\in\mathcal R_r$ has a column
space not contained in $\operatorname{col}(A)$.
Thus neither class contains the other.

Both classes contain zero, so their sum contains
their union. Since $X$ and $Y$ have orthogonal
ranges and are positive semidefinite,
$\operatorname{rank}(X+Y)=N+1$.
This exceeds the maximum rank of either class,
proving strict inclusion.

Finally, rank subadditivity gives
\[
  \operatorname{rank}(X'+Y')
  \le
  \operatorname{rank}(X')+\operatorname{rank}(Y')
  \le N+r
\]
for every $X'\in\mathcal I_A$ and
$Y'\in\mathcal R_r$.
Together with the output dimension $M$,
this proves~\eqref{eq:dune-dual-rank}.
\end{proof}

\paragraph{Implications for dual-resolution design.}
The auxiliary branch compensates for the limited
representation of the rank-$r$ branch while imposing
its own frequency constraint.
Conversely, the SVD branch provides adaptive singular
directions unavailable within the fixed interpolation
space.
This complementarity motivates joint optimization
rather than relying on either representation alone.

\begin{remark}[Scope Under Subsequent Processing]
The representation classes characterize the outputs
of SVD truncation and bilinear interpolation,
respectively. Separable Gaussian smoothing does not
increase matrix rank, whereas entrywise clipping
and masking can alter both rank and frequency content.
Consequently, the rank and HF bounds established
above do not necessarily hold for the final effective
perturbation $\eta=x_p-x$.
\end{remark}

\section{Rank Estimates of PGD Perturbations}
\label{sec:supp-null}
We set the fixed SVD rank using five 100-source CelebA-HQ samples (seeds 0--4), with identities disjoint from the evaluation set and from one another.
Plain PGD perturbations use $\epsilon=16/255$, $\alpha=1/255$, and 100 normalized gradient-direction steps against the ensemble of the three non-held-out recognizers, with no mask, grid, SVD truncation, smoothing, or TV.
We apply~\eqref{eq:dune-rmt-rank} to each $256\times256$
color channel of the effective perturbation $\eta=x_p-x$.
Each 100-source sample therefore yields
$100\times4\times3=1{,}200$ channel-wise rank estimates.
As summarized in \Cref{tab:supp-rank-seeds}, the five samples give median ranks of 10 or 11, with a pooled median of 11, which we use throughout optimization.
The evaluation-set estimates are included in the table for reference.

\begin{table}[!ht]
\centering
\caption{\textbf{GD rank estimates of plain PGD perturbations.}
$\widehat{r}$ from~\eqref{eq:dune-rmt-rank} per color channel;
$\epsilon=16/255$, 100 gradient-direction steps, four held-out
recognizers. Seeds 0--4 each sample 100 CelebA-HQ sources
(one per identity) whose identities are disjoint from the
evaluation set and from the other seeds, 500 identities in total. Energy is the mean share of
$\sum_i s_i^2$ in the retained components.}
\label{tab:supp-rank-seeds}
\small
\begin{tabular}{lccc}
\toprule
Sample & Median [IQR] & Mean & Energy \\
\midrule
Seed 0 & 11 [9, 13] & 11.04 & 27.6\% \\
Seed 1 & 11 [9, 13] & 11.33 & 28.3\% \\
Seed 2 & 11 [9, 13] & 10.97 & 27.2\% \\
Seed 3 & 10 [8, 13] & 10.86 & 27.0\% \\
Seed 4 & 11 [9, 13] & 11.14 & 27.7\% \\
\midrule
Seeds 0--4 & 11 [9, 13] & 11.07 & 27.5\% \\
\midrule
Evaluation set & 11 [9, 13] & 11.15 & 28\% \\
\bottomrule
\end{tabular}
\end{table}

\section{Singular-Component Frequency Profiles at $256\times256$}
\label{sec:supp-component-256}
\begin{figure*}[t]
\centering
\includegraphics[width=\textwidth]{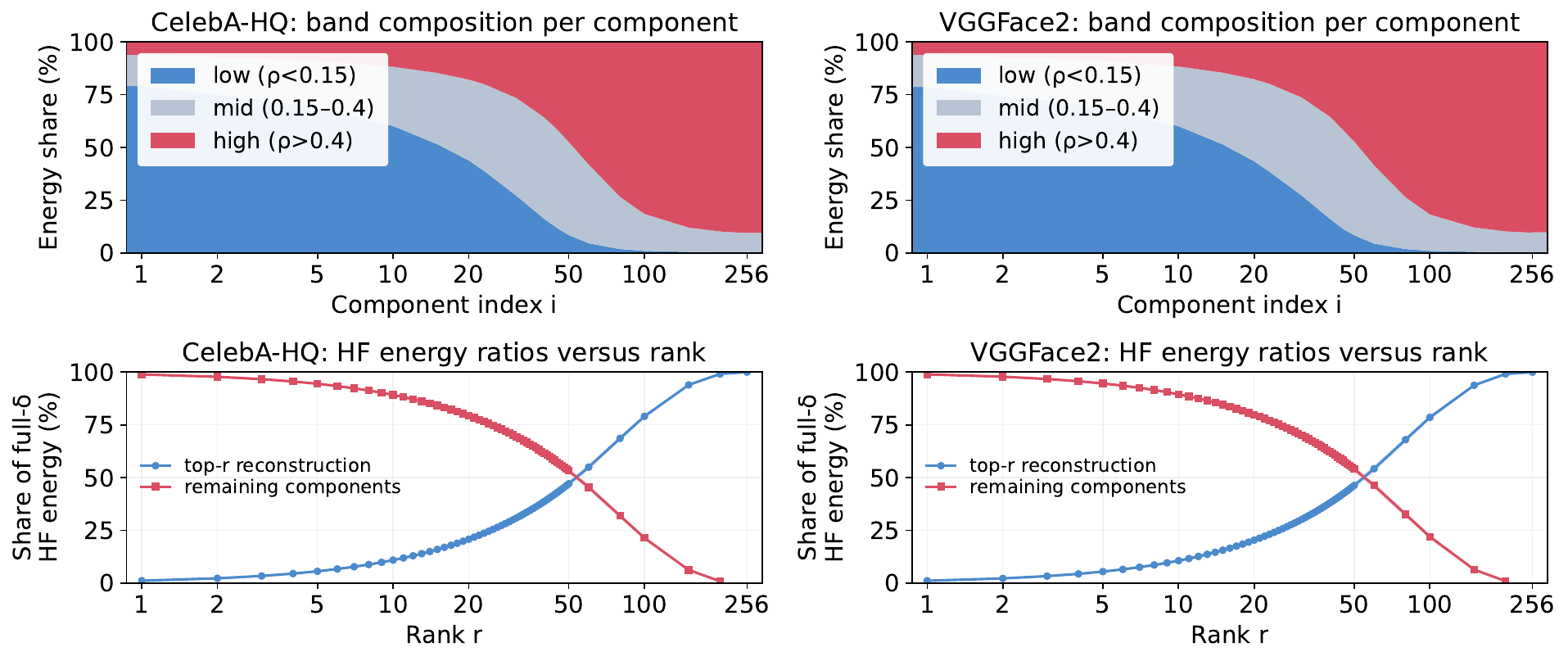}
\caption{\textbf{Singular-component frequency profiles at $256\times256$.} PGD perturbations on 200 sources each from the CelebA-HQ (left) and VGGFace2 (right) evaluation sets, with the same settings as main-paper \Cref{fig:component}. Top: frequency-band energy shares per singular component. Bottom: HF energy ($\rho>0.4$) of the top-$r$ reconstruction and the remaining components, normalized by the full perturbation's HF energy and averaged over images.}
\label{fig:supp-component-256}
\end{figure*}

Main-paper \Cref{fig:component} analyzes PGD perturbations on 1{,}000 CelebA-HQ images at $512\times512$.
To examine the same pattern at the resolution used in the main experiments and on another dataset, \Cref{fig:supp-component-256} repeats the analysis at $256\times256$ on the first 200 sources from each of the CelebA-HQ and VGGFace2 evaluation sets.
The attack settings and frequency bands are the same as in \Cref{fig:component}.
Both datasets show a similar trend: leading components have a larger low-frequency share, while the high-frequency share increases toward later components.
At rank 11, the remaining components carry 88.2\% (CelebA-HQ) and 88.5\% (VGGFace2) of the full perturbation's HF energy, compared with 88.4\% at $512\times512$.
The observations at $256\times256$ are thus consistent across the two datasets and with the $512\times512$ analysis.

\section{Evaluation Protocol and Detailed Results}
\label{sec:supp-protocol}
The VGGFace2 evaluation set comprises 500 pairs sampled from 500 distinct identities, each pair containing two different images of the same identity, aligned with the same dlib/FFHQ procedure.

For face-swap verification, the query is a generated swap and the reference is the original clean source.
For dodging, the query is the protected source and the reference is a second same-identity portrait.
The cosine similarities in \Cref{tab:supp-transform-cos,tab:supp-purify-cos} use this same paired clean reference portrait, a different image of the same identity; the Clean rows use the unprotected source as the query, with the indicated processing applied.
Retrieval uses all 500 clean reference portraits; R1-P/R5-P count at least one/five candidates with strictly higher similarity than the paired reference.
Landmark failures count as protection failures and remain in the denominator.

Face-swap and dodging scoring use thresholds $(0.241,0.167,0.409,0.302)$ for IR-SE50, IR152, FaceNet, and MobileFace.
These are inherited nominal FAR-0.01 constants; on our CelebA-HQ set, their measured impostor FAR over all 249{,}500 non-matching probe--reference pairs is 1.05\%, 1.38\%, 0.84\%, and 0.39\%, respectively.
The surrogate landmark warp is fixed during protection; evaluation detects landmarks on the protected/generated query.
Face-Adapter uses Stable Diffusion 1.5, 25 Euler steps, and $512^2$ swap generation.
None of the four swap generators is used for protection optimization.
For the transformations in \Cref{fig:robust-transforms}, JPEG uses 4:2:0 chroma subsampling, and both protected and clean query controls are transformed while the reference gallery and recognition thresholds stay fixed.

\subsection{Protection Generation Settings}
\label{sec:supp-generation-settings}
We report the configurations actually used for the main comparisons in
\Cref{tab:supp-generation-settings}.
Published method configurations provide the starting point; the following
task adaptations and experiment-specific iteration budgets are explicit.
All methods use the same evaluation pairs and leave-one-out recognizer splits.
Generative methods are not projected onto a common additive perturbation bound.

\begin{table*}[t]
\centering\footnotesize
\caption{\textbf{Generation settings for all methods in the main comparisons.} For noise-based methods, bounds and step coefficients are in units of $1/255$; normalized-gradient and sign updates differ. Generative methods use their method-specific settings below and are not projected onto a common additive perturbation bound. AMT-GAN and DiffAM use the source-identity loss adaptations described in the text.}
\label{tab:supp-generation-settings}
\setlength{\tabcolsep}{5pt}
\begin{tabular*}{\textwidth}{@{\extracolsep{\fill}}lcccp{0.50\textwidth}}
\toprule
\multicolumn{5}{l}{\textbf{Noise-based methods and DUNE}} \\
\midrule
Method & $\epsilon$ & Iterations & $\alpha$ & Update and method-specific parameters \\
\midrule
PGD & 16 & 10 & 1.6 & Sign updates; uniform random start in $[-\epsilon,\epsilon]$. \\
MI-FGSM & 16 & 10 & 1.6 & Momentum sign updates, $\mu=1$; the same random-start bound. \\
TI-DIM & 16 & 10 & 1.6 & Momentum $\mu=1$; $15\times15$ TI kernel, $\sigma=3$; input-diversity probability $0.5$, resize lower ratio $0.875$; the same random-start bound. \\
TIP-IM & 12 & 50 & 0.36 & L2-normalized gradient, step factor $1.5$, ten targets, MMD weight $\gamma=5\times10^{-4}$. \\
DUNE (Ours) & 16 & 100 & 1.0 & Main branch: normalized gradient, $\epsilon_r=16$, rank 11, $3\times3$ Gaussian with $\sigma_r=0.4$. Grid: $96\times96$, sign updates, $\epsilon_g=8$. Importance: one mask on the summed perturbation, 20 probes, top 30\%; masked TV weight 20. \\
\bottomrule
\end{tabular*}

\vspace{6pt}
\begin{tabular*}{\textwidth}{@{\extracolsep{\fill}}lp{0.80\textwidth}}
\toprule
\multicolumn{2}{l}{\textbf{Makeup-based and diffusion-based methods}} \\
\midrule
Method & Configuration \\
\midrule
AMT-GAN~\cite{amtgan} & Five training epochs, batch size 1; generator and discriminator learning rates $2\times10^{-4}$, Adam betas $(0.5,0.999)$. Loss weights: GAN 10, cycle 10, adversarial 5, makeup 2, identity 5; histogram weights for lips/skin/eyes: $1/0.1/1$. Train a separate generator for each held-out recognizer. \\
DiffAM~\cite{diffam} & Adam learning rate $4\times10^{-6}$, multiplier $1.2$ per epoch, six epochs, 200 training images; diffusion time $t_0=60$, 20 inversion steps and six sampling steps during training and editing. Reference makeup XMY-060; four target-specific models per held-out recognizer, with target portraits 005869/085807/116481/169284 assigned to four source groups. \\
CLIP2Protect~\cite{clip2protect} & Generator tuning: 450 iterations, learning rate $5\times10^{-4}$, L2 weight $0.5$. Adversarial latent optimization: 50 Adam iterations, learning rate $0.01$; weights $(\lambda_{\rm adv},\lambda_{\rm CLIP},\lambda_{\rm latent})=(1,0.5,0.01)$; source text ``face'', makeup prompt ``red lipstick'', one augmentation, and noise optimization enabled. Four target groups; seed 0. \\
DiffProtect~\cite{diffprotect} & DiffAE FFHQ-256 backbone; 50 attack iterations, latent bound $\gamma=0.03$, step size $2\gamma/50=0.0012$, parsing regularization $\lambda=0$; 100 encoding/final decoding steps and five attack-time decoding steps, starting at step 5; untargeted identity loss. \\
WDP~\cite{diffpurify} & Stable Diffusion 2-base; 20 DDIM inversion steps, reverse-process start index 17; 20 unconditional-embedding AdamW iterations at learning rate $0.1$, 35 adversarial-latent AdamW iterations at learning rate $0.01$, adversarial weight $0.003$; obfuscation mode. \\
\bottomrule
\end{tabular*}
\end{table*}

The PGD, MI-FGSM, and TI-DIM rows use the ten-iteration outputs in both
\Cref{tab:dodging,tab:faceswap}; the main DUNE row uses 100 iterations.
Thus these main-table rows do not constitute an iteration-matched comparison.
The separate comparison in \Cref{sec:budget-tradeoff} uses ten iterations,
$\epsilon=16/255$, and $\alpha=1.6/255$ throughout. TIP-IM there uses sign
updates ($\mathrm{norm}=\ell_\infty$), step factor 1, ten targets, and
$\gamma=5\times10^{-4}$; DUNE retains its normalized-gradient main branch
and sign-updated grid.

For AMT-GAN and DiffAM, the main results use models retrained with a
source-identity departure term added to their target-attraction objective.
Writing $c_t$ and $c_s$ for cosine similarity of the generated face to the
target and clean source, respectively, the adversarial term is
$(1-c_t)-(1-c_s)$, with equal coefficients on attraction and departure.
Other training settings are retained as listed above. This adaptation
supports dodging while retaining each method's makeup-generation mechanism.
The other methods use their implemented dodging/obfuscation modes.
Generation-time recognition losses use a fixed clean-source landmark warp;
query landmarks are detected independently during evaluation.

\begin{table*}[!t]
\centering\footnotesize
\begin{tabular}{lp{0.80\textwidth}}
\toprule
Method & Configuration \\
\midrule
FaceShield~\cite{jeong2025faceshield} &
$\epsilon=16/255$, 100 sign-PGD iterations, and step size $1/255$, matching our method.
The official losses (face detection, identity, and Stable Diffusion UNet losses) and their weights are kept, as is the official per-step filtering (Gaussian smoothing, block-DCT low-pass filtering, and $L_\infty$ clipping).
Task adaptation: the ArcFace identity loss is replaced by the cosine loss averaged over the three surrogate recognizers.
Inputs are processed at $256\times256$. \\
\midrule
NullSwap$^{*}$~\cite{wang2025nullswap} &
Generator architecture and losses (identity, MSE, LPIPS, and adversarial) follow the paper, including its loss weights and dynamic loss weighting; no $L_\infty$ bound on the output.
Trained for 60 epochs at $256\times256$ on 21{,}390 CelebA-HQ images whose identities do not overlap with the evaluation set; the same generators are applied to VGGFace2.
Task adaptation: one generator per held-out recognizer, trained with the three surrogates.
Due to the memory limits of our GPUs, we use a batch size of 32 instead of the 256 used in the paper. \\
\bottomrule
\end{tabular}
\caption{\textbf{Generation settings for the face-swap defense methods.} $^{*}$Reimplemented from the paper, as no official code is available.}
\label{tab:supp-faceswap-defense-settings}
\end{table*}

\begin{table*}[!t]
\centering
\caption{\textbf{Comparison with face-swap defense methods.} Dodging PSR, R1-P, R5-P, and face-swap PSR (\%) average four held-out recognizers; Avg.\ averages the four swappers. $^{*}$Reimplemented from the paper.}
\label{tab:supp-faceswap-defense}
\footnotesize
\setlength{\tabcolsep}{3.2pt}
\resizebox{\textwidth}{!}{%
\begin{tabular}{ll|ccc|ccccc|cccc}
\toprule
& & \multicolumn{3}{c|}{Dodging} & \multicolumn{5}{c|}{Face-swap PSR$\uparrow$} & \multicolumn{4}{c}{Source fidelity} \\
Dataset & Method & PSR$\uparrow$ & R1-P$\uparrow$ & R5-P$\uparrow$ & SimSwap & INSw. & UniFace & FA & Avg. & PSNR$\uparrow$ & SSIM$\uparrow$ & FID$\downarrow$ & LPIPS$\downarrow$ \\
\midrule
\multirow{3}{*}{CelebA-HQ}
& FaceShield~\cite{jeong2025faceshield} & \second{84.20} & \second{89.60} & \second{81.15} & \second{94.3} & \second{69.3} & \second{89.2} & \second{83.3} & \second{84.0} & 28.89 & 0.850 & 78.6 & 0.143 \\
& NullSwap$^{*}$~\cite{wang2025nullswap} & 55.25 & 68.50 & 52.65 & 60.2 & 45.0 & 58.1 & 64.4 & 56.9 & \second{30.23} & \second{0.943} & \best{8.4} & \best{0.018} \\
& Ours & \best{93.80} & \best{96.70} & \best{92.85} & \best{98.4} & \best{88.4} & \best{98.5} & \best{94.9} & \best{95.1} & \best{33.20} & \best{0.957} & \second{28.0} & \second{0.035} \\
\midrule
\multirow{3}{*}{VGGFace2}
& FaceShield~\cite{jeong2025faceshield} & \second{78.55} & \second{85.65} & \second{74.10} & \second{88.9} & \second{65.3} & \second{92.7} & \second{83.6} & \second{82.6} & 28.82 & 0.836 & 86.6 & 0.207 \\
& NullSwap$^{*}$~\cite{wang2025nullswap} & 47.35 & 57.55 & 39.45 & 52.8 & 38.2 & 65.1 & 61.7 & 54.4 & \second{30.19} & \second{0.937} & \best{13.0} & \best{0.034} \\
& Ours & \best{90.65} & \best{94.00} & \best{88.35} & \best{95.6} & \best{83.5} & \best{99.3} & \best{94.0} & \best{93.1} & \best{33.29} & \best{0.953} & \second{24.0} & \second{0.058} \\
\bottomrule
\end{tabular}}
\end{table*}
\paragraph{Face-swap defense methods.}
\Cref{tab:supp-faceswap-defense-settings} lists the settings for the two
face-swap defenses, FaceShield~\cite{jeong2025faceshield} and
NullSwap~\cite{wang2025nullswap}. Both use the leave-one-out recognizer
splits of the main comparisons: their identity losses are computed with the
three surrogate recognizers in place of the recognizers used in the original
papers. FaceShield keeps the losses and filtering of its paper but runs with the $L_\infty$ budget and iteration count of our method ($\epsilon=16/255$, 100 iterations); loss weights and thresholds that the paper does not report are taken from the official implementation.
NullSwap has no public implementation; we reimplement it from the paper,
keep all reported hyperparameters, and list the unreported details we fix.

\paragraph{Comparison with face-swap defenses.}
\Cref{tab:supp-faceswap-defense} compares both defenses with our method on the evaluation sets, recognizers, and swappers of the main comparisons.
Our method gives the highest dodging and face-swap PSR in every column on both datasets.
At the same budget and iteration count, FaceShield is 9.6 and 12.1 points lower in dodging PSR on CelebA-HQ and VGGFace2, with lower source fidelity (PSNR 28.9, FID 78.6 on CelebA-HQ).
FaceShield applies DCT low-pass filtering during optimization; the controlled comparison in main-paper \Cref{tab:refinement} also shows lower source fidelity for DCT filtering than for SVDG.
NullSwap$^{*}$ attains the lowest FID and LPIPS, but its outputs are not $L_\infty$-bounded (about $100/255$), and its generator protects 55.3\% and 47.4\% of the sources under dodging.

\begin{figure*}[t]
\centering
\includegraphics[width=\linewidth]{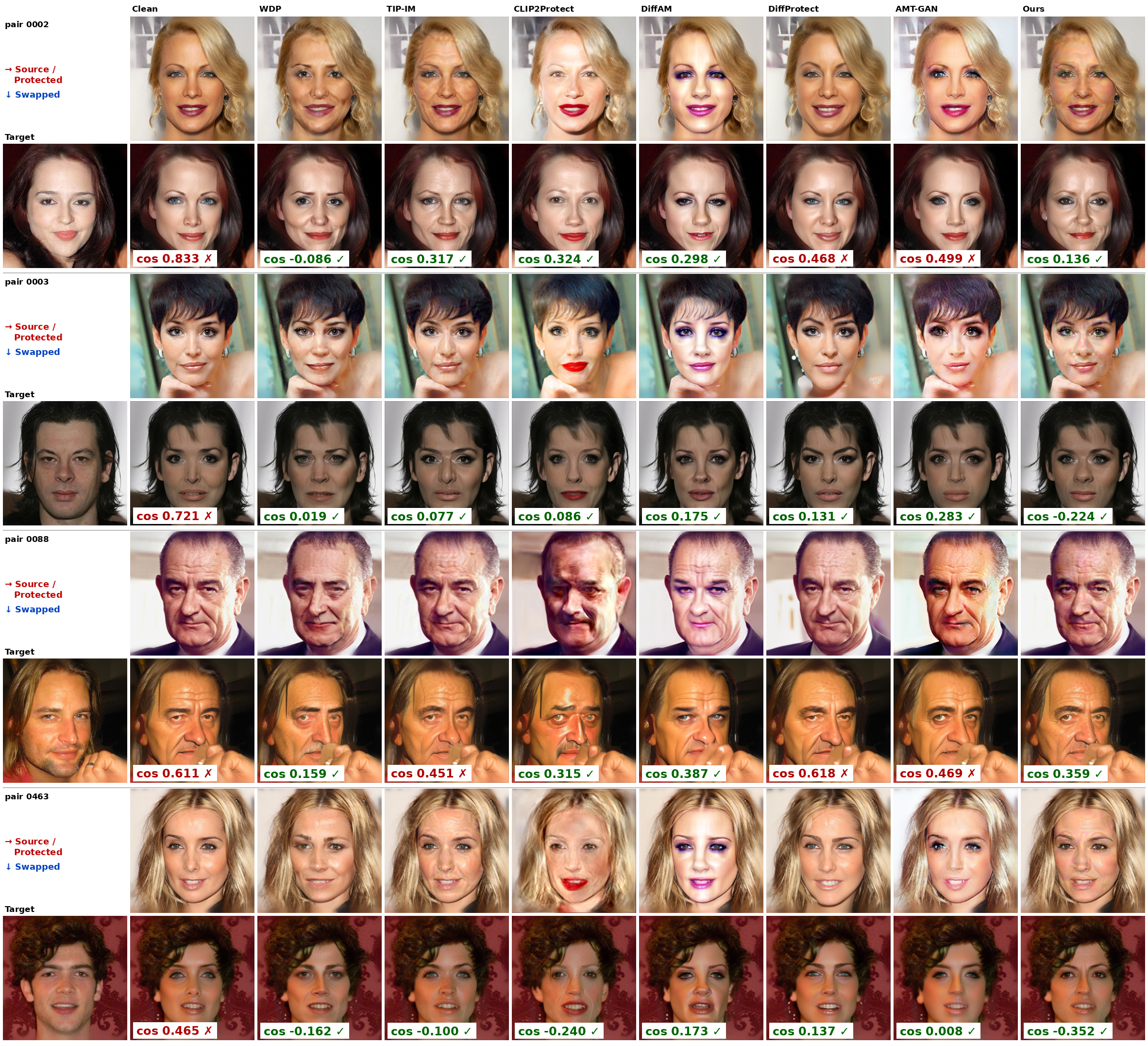}
\caption{For each pair, the top row shows the clean or protected source and the bottom row the swap output, with the target at the left.
Labels give output-to-clean-source cosine similarity; green checks mark values below the FaceNet threshold $\tau=0.409$.}
\label{fig:supp-qual-a}
\end{figure*}

\begin{figure*}[t]
\centering
\includegraphics[width=\linewidth]{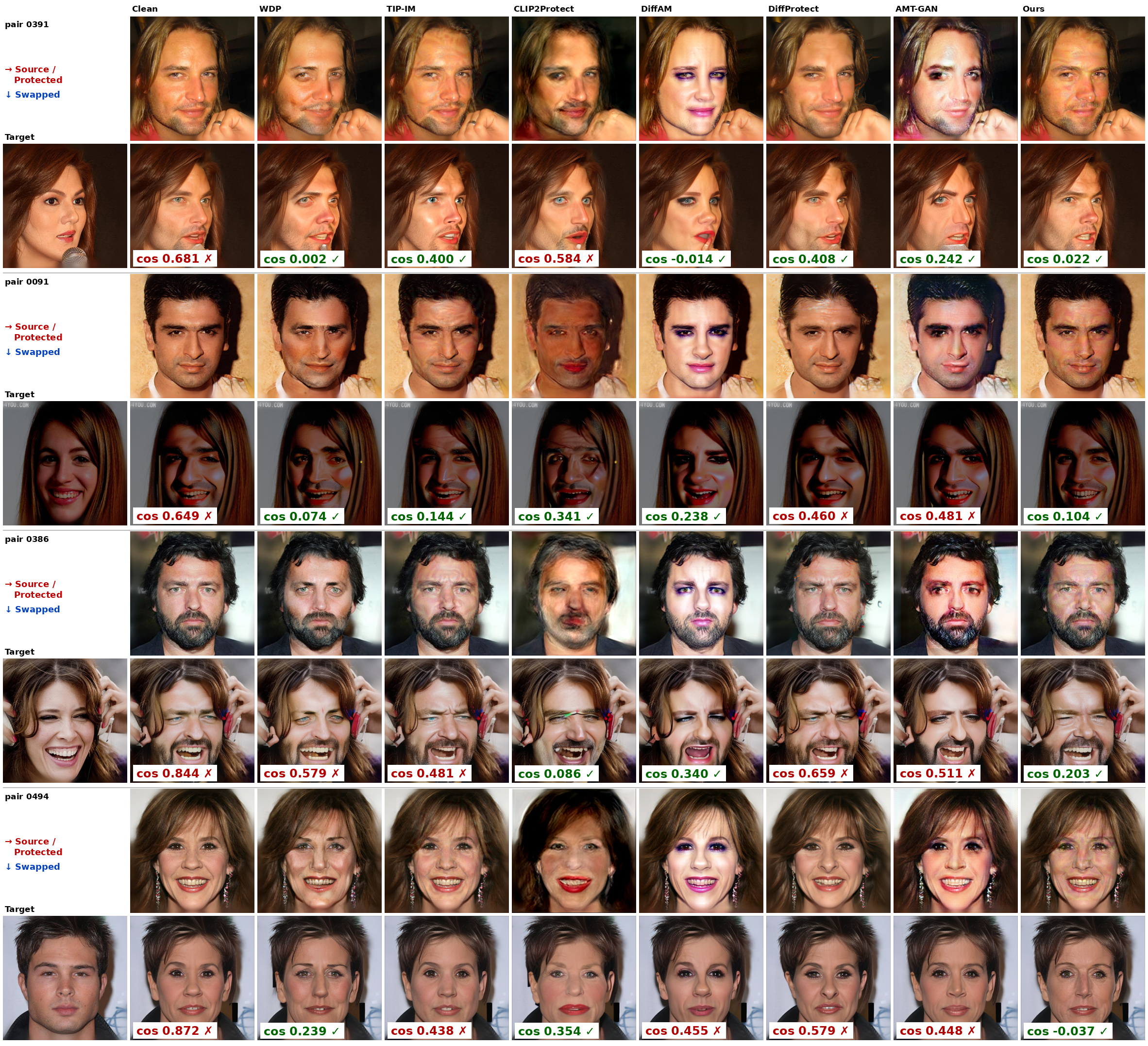}
\caption{\textbf{Additional Face-Adapter examples on CelebA-HQ}
For each pair, the top row shows the clean or protected source and the bottom row the swap output, with the target at the left.
Labels give output-to-clean-source cosine similarity; green checks mark values below the FaceNet threshold $\tau=0.409$.}
\label{fig:supp-qual-b}
\end{figure*}

\paragraph{Additional examples.}
\Cref{fig:supp-qual-a,fig:supp-qual-b} show eight additional Face-Adapter pairs on CelebA-HQ (none is the main-paper example).
The held-out evaluator is FaceNet and TIP-IM uses its original settings.


\begin{table*}[!t]
\centering\small
\caption{\textbf{Budget-matched dodging protection on CelebA-HQ and VGGFace2.} All methods use $\epsilon=16/255$, ten iterations, and step coefficient $\alpha=1.6/255$; update directions follow \Cref{sec:supp-generation-settings}. PSR/R1-P/R5-P (\%) average four held-out recognizers. CelebA-HQ face-swap results are in main-paper \Cref{tab:budget-faceswap}.}
\label{tab:budget}
\setlength{\tabcolsep}{4pt}
\begin{tabular*}{\textwidth}{@{\extracolsep{\fill}}ll*{8}{c}@{}}
\toprule
Dataset & Method & PSR$\uparrow$ & R1-P$\uparrow$ & R5-P$\uparrow$ & PSNR$\uparrow$ & SSIM$\uparrow$ & FID$\downarrow$ & LPIPS$\downarrow$ & $|\delta|\downarrow$ \\
\midrule
\multirow{6}{*}{CelebA-HQ} & Clean & 9.00 & 11.95 & 8.40 & -- & -- & -- & -- & -- \\
 & PGD & 70.00 & 80.00 & 66.40 & 28.31 & 0.668 & 61.1 & 0.194 & 8.39 \\
 & MI-FGSM & 75.20 & 82.65 & 69.70 & 27.59 & 0.644 & 80.5 & 0.233 & 9.23 \\
 & TI-DIM & 66.65 & 77.35 & 59.65 & 27.29 & 0.690 & 85.2 & 0.191 & 9.61 \\
 & TIP-IM & \second{79.40} & \second{86.75} & \second{75.20} & \second{30.06} & \second{0.846} & \second{60.3} & \second{0.118} & \second{5.34} \\
 & Ours & \best{88.70} & \best{92.85} & \best{86.60} & \best{35.44} & \best{0.964} & \best{19.6} & \best{0.024} & \best{1.91} \\
\midrule
\multirow{6}{*}{VGGFace2} & Clean & 0.80 & 1.20 & 0.40 & -- & -- & -- & -- & -- \\
 & PGD & 60.85 & 71.10 & 55.50 & 28.33 & 0.649 & 94.7 & 0.316 & 8.37 \\
 & MI-FGSM & 68.00 & 76.15 & 62.10 & 27.58 & 0.622 & 109.9 & 0.353 & 9.23 \\
 & TI-DIM & 62.20 & 69.65 & 54.95 & 27.28 & 0.675 & 119.9 & 0.285 & 9.62 \\
 & TIP-IM & \second{75.35} & \second{83.65} & \second{71.20} & \second{30.13} & \second{0.838} & \second{57.1} & \second{0.200} & \second{5.32} \\
 & Ours & \best{82.90} & \best{89.05} & \best{79.95} & \best{35.54} & \best{0.961} & \best{18.9} & \best{0.043} & \best{1.89} \\
\bottomrule
\end{tabular*}
\end{table*}

\begin{figure*}[!t]
\centering
\includegraphics[width=0.9\linewidth]{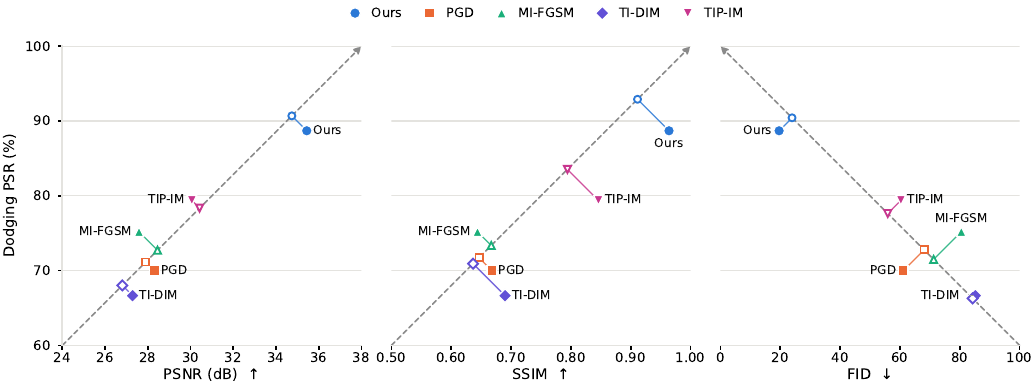}
\caption{\textbf{Joint comparison of attack performance and image quality on CelebA-HQ.} Dodging attack performance (PSR) is plotted against image-quality metrics (PSNR, SSIM, and FID) with $\epsilon=16/255$, 10 iterations, and $\alpha=1.6/255$, using the results in \Cref{tab:budget}. Filled markers show the measured results for each method. Dashed diagonals indicate the direction of jointly improving attack performance and image quality; open markers are geometric projections onto these guides in normalized panel coordinates.}
\label{fig:visualpsr}
\end{figure*}
\section{Comparison Under the Same Budget}
\label{sec:isobudget}

\paragraph{Ten-iteration comparison on CelebA-HQ.}
\Cref{tab:budget} compares dodging protection and source fidelity on both datasets under the matched settings of main-paper \Cref{tab:budget-faceswap}.

\begin{figure*}[!t]
\centering
\includegraphics[width=\linewidth]{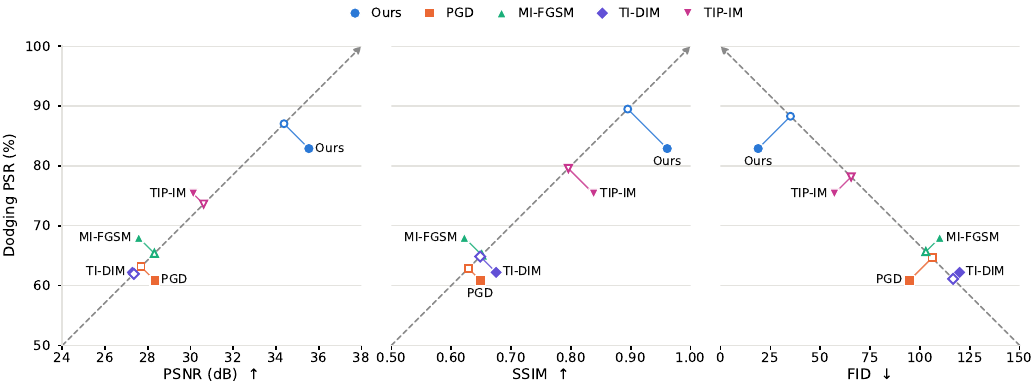}
\caption{\textbf{Joint comparison of attack performance and image quality on VGGFace2.} Dodging attack performance (PSR) is plotted against image-quality metrics (PSNR, SSIM, and FID) with $\epsilon=16/255$, 10 iterations, and $\alpha=1.6/255$, using the results in \Cref{tab:budget}. Filled markers show the measured results for each method. Dashed diagonals indicate the direction of jointly improving attack performance and image quality; open markers are geometric projections onto these guides in normalized panel coordinates.}
\label{fig:supp-visualpsr-vggface2}
\end{figure*}

\Cref{fig:visualpsr} displays these protection and quality outcomes jointly.
The same ordering holds on VGGFace2 (\Cref{tab:budget,fig:supp-visualpsr-vggface2}).

\begin{table*}[!t]
\centering
\caption{\textbf{Effect of the number of iterations.} Complete method with $\epsilon=16/255$ and $\alpha=1/255$; only the iteration count changes. Dodging and face-swap PSR (\%) average four held-out recognizers; Avg.\ averages the four swappers. The 10-iteration setting of \Cref{tab:budget} instead uses $\alpha=1.6/255$.}
\label{tab:supp-iterations}
\footnotesize
\setlength{\tabcolsep}{4pt}
\begin{tabular*}{\textwidth}{@{\extracolsep{\fill}}llc|ccccc|ccc@{}}
\toprule
& & Dodging & \multicolumn{5}{c|}{Face-swap PSR$\uparrow$} & \multicolumn{3}{c}{Source fidelity} \\
Dataset & Iterations & PSR$\uparrow$ & SimSwap & INSw. & UniFace & FA & Avg. & PSNR$\uparrow$ & SSIM$\uparrow$ & FID$\downarrow$ \\
\midrule
\multirow{5}{*}{CelebA-HQ}
& 10 & 81.45 & 84.5 & 70.4 & 85.7 & 83.3 & 81.0 & 37.02 & 0.973 & 15.3 \\
& 30  & 92.80 & 97.4 & 87.4 & 96.9 & 94.0 & 93.9 & 34.44 & 0.962 & 21.1 \\
& 50  & 93.60 & 98.2 & 88.5 & 98.1 & 94.7 & 94.9 & 33.81 & 0.960 & 24.0 \\
& 70 & 93.40 & 98.7 & 88.7 & 98.5 & 95.0 & 95.2 & 33.48 & 0.958 & 25.8 \\
& 100 & 93.80 & 98.4 & 88.4 & 98.5 & 94.9 & 95.1 & 33.20 & 0.957 & 28.0 \\
\midrule
\multirow{5}{*}{VGGFace2}
& 10 & 74.15 & 77.1 & 63.5 & 89.3 & 80.5 & 77.6 & 37.13 & 0.971 & 15.6 \\
& 30  & 88.60 & 93.8 & 81.6 & 98.6 & 91.5 & 91.4 & 34.55 & 0.959 & 19.6 \\
& 50 & 89.25 & 95.0 & 83.1 & 98.9 & 92.8 & 92.5 & 33.91 & 0.956 & 21.4 \\
& 70 & 89.80 & 95.0 & 82.8 & 99.0 & 93.9 & 92.7 & 33.58 & 0.954 & 22.6 \\
& 100 & 90.65 & 95.6 & 83.5 & 99.3 & 94.0 & 93.1 & 33.29 & 0.953 & 24.0 \\
\bottomrule
\end{tabular*}
\end{table*}
\paragraph{Ten-iteration comparison on VGGFace2.}
The VGGFace2 block of \Cref{tab:budget} uses the same update settings as the CelebA-HQ block.
Our method achieves the highest PSR ($82.90\%$ versus $75.35\%$ for TIP-IM), R1-P, and R5-P, together with the highest PSNR/SSIM and lowest FID/LPIPS.
\Cref{fig:supp-visualpsr-vggface2} jointly compares attack performance and image quality on VGGFace2.

\section{Effect of the Number of Iterations}
\label{sec:supp-iterations}
We use 100 iterations in the main comparisons to allow protection to develop under a common optimization budget on both datasets.
\Cref{tab:supp-iterations} varies only the number of optimization steps of the complete method, keeping $\epsilon=16/255$, $\alpha=1/255$, and every other setting of the main configuration.
The 10-, 30-, 50-, and 70-step outputs are snapshots of a single 100-step run, which equal separate runs with fewer steps because the step size is fixed; the 100-step rows are the main results.
Dodging PSR, face-swap PSR, and source fidelity are measured as in the main paper.
At 30 iterations, mean face-swap PSR already reaches 93.9\% on CelebA-HQ and 91.4\% on VGGFace2, compared with 95.1\% and 93.1\% at 100 iterations.
The shorter runs retain higher PSNR and lower FID, showing the protection--fidelity trade-off along optimization.

\section{Role of the Low-Resolution Branch}
\label{sec:supp-joint}
On its own, the low-resolution (auxiliary) branch is a weak attack. Optimized alone, it reaches only 55.90\% dodging PSR (\Cref{tab:supp-joint}), because bilinear upsampling restricts its perturbation to smooth, low-frequency patterns (\Cref{sec:suppl_frequency_bilinear}).
In DUNE, this branch is optimized jointly with the main branch and supplements the main-branch perturbation with additional low-frequency components. This raises dodging PSR from 90.55\% with the main branch alone to 93.80\%, with better image quality.
Joint optimization thus makes effective use of the low-resolution branch, which is of limited use on its own.
\begin{table}[!htbp]
\centering\footnotesize
\caption{\textbf{Role of the low-resolution branch on CelebA-HQ.} Auxiliary only and Main only use no masking or TV; Ours is the full method. Dodging PSR (\%) averages four held-out recognizers. Bold/underline: best/second best.}
\label{tab:supp-joint}
\fontsize{9}{10}\selectfont
\renewcommand{\arraystretch}{0.95}
\setlength{\tabcolsep}{3pt}
\begin{tabular*}{\columnwidth}{@{\extracolsep{\fill}}lccccc@{}}
\toprule
Method & PSR$\uparrow$ & PSNR$\uparrow$ & SSIM$\uparrow$ & LPIPS$\downarrow$ & FID$\downarrow$ \\
\midrule
Auxiliary only & 55.90 & \textbf{34.05} & 0.909 & 0.070 & 38.3 \\
Main only & \second{90.55} & 32.77 & \second{0.926} & \second{0.059} & \second{30.2} \\
Ours & \textbf{93.80} & \second{33.20} & \textbf{0.957} & \textbf{0.035} & \textbf{28.0} \\
\bottomrule
\end{tabular*}
\end{table}

\begin{figure*}[!t]
\centering
\includegraphics[width=\linewidth]{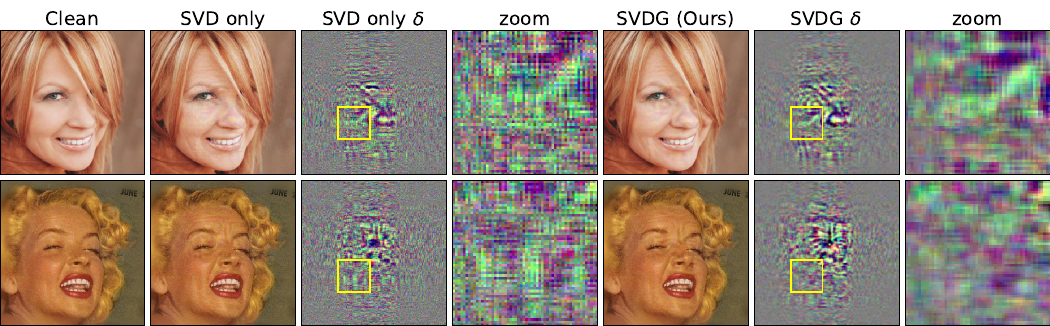}
\caption{\textbf{Artifacts of SVD truncation and their mitigation by Gaussian smoothing.} CelebA-HQ examples after 10 iterations. $\delta$ columns show the amplified perturbation; zoom columns enlarge the yellow box with nearest-neighbor upsampling. SVD only: rank 11 without smoothing; SVDG: rank 11 followed by the Gaussian.}
\label{fig:supp-svdg-artifact}
\end{figure*}
\section{Artifacts of SVD Truncation}
\label{sec:supp-svd-gauss}
The quantitative comparison of SVD truncation, Gaussian smoothing, and their combination is given in main-paper \Cref{tab:refinement}.

\Cref{fig:supp-svdg-artifact} illustrates the effect of Gaussian smoothing after SVD truncation. In these examples, the amplified SVD-only perturbations exhibit horizontal and vertical streaks and block-like patterns. Applying the $3\times3$ Gaussian ($\sigma=0.4$) after truncation at each step reduces these visible artifacts.

\section{Post-hoc versus Per-Step SVD}
\label{sec:supp-svd-timing}
We compare truncating the perturbation once after PGD optimization with truncating it at every step during optimization, as in DUNE (\Cref{tab:supp-svd-timing}). Both use rank 11 per channel. All methods are run under identical conditions. For post-hoc SVD, the final PGD perturbation is truncated to rank 11 once, without further optimization.
\begin{table}[!htbp]
\centering\footnotesize
\caption{\textbf{Post-hoc versus per-step SVD (rank 11) on CelebA-HQ.} Dodging PSR (\%) averages four held-out recognizers. Bold/underline: best/second best.}
\label{tab:supp-svd-timing}
\setlength{\tabcolsep}{3pt}
\resizebox{\columnwidth}{!}{%
\begin{tabular}{lccccc}
\toprule
Condition & PSR$\uparrow$ & PSNR$\uparrow$ & SSIM$\uparrow$ & LPIPS$\downarrow$ & FID$\downarrow$ \\
\midrule
PGD & \best{86.10} & 27.78 & 0.652 & 0.2160 & 70.2 \\
Post-hoc SVD & 46.80 & \best{33.61} & \best{0.905} & \best{0.0553} & \best{34.1} \\
\midrule
Per-step SVD & \second{85.50} & \second{31.47} & \second{0.854} & \second{0.0975} & \second{45.3} \\
\bottomrule
\end{tabular}}
\end{table}
A single truncation after optimization reduces PSR from 86.10\% to 46.80\%, whereas per-step truncation retains 85.50\%. Per-step truncation limits each channel to rank at most 11 at the SVD stage of every iteration, allowing subsequent updates to adapt to the truncation.

\begin{table*}[!t]
\centering\footnotesize
\caption{\textbf{Identity similarity after image transformations on CelebA-HQ.} Mean cosine similarity between each query image and the dodging reference, a different unmodified portrait of the same identity, over four held-out recognizers (lower is better); Clean denotes the unprotected source images, at the levels of \Cref{fig:robust-transforms}. Bold/underline: best/second best.}
\label{tab:supp-transform-cos}
\setlength{\tabcolsep}{4pt}
\begin{tabular}{@{}p{0.16\textwidth}|*{1}{>{\centering\arraybackslash}p{\dimexpr(\textwidth-0.16\textwidth-20\tabcolsep-4\arrayrulewidth)/10\relax}}|*{3}{>{\centering\arraybackslash}p{\dimexpr(\textwidth-0.16\textwidth-20\tabcolsep-4\arrayrulewidth)/10\relax}}|*{3}{>{\centering\arraybackslash}p{\dimexpr(\textwidth-0.16\textwidth-20\tabcolsep-4\arrayrulewidth)/10\relax}}|*{3}{>{\centering\arraybackslash}p{\dimexpr(\textwidth-0.16\textwidth-20\tabcolsep-4\arrayrulewidth)/10\relax}}@{}}
\toprule
 & & \multicolumn{3}{c|}{JPEG quality (\%)} & \multicolumn{3}{c|}{Resize (\%)} & \multicolumn{3}{c}{Blur ($\sigma$)} \\
Method & None & 75 & 50 & 10 & 75 & 50 & 25 & 1 & 2 & 3 \\
\midrule
Clean & 0.597 & 0.596 & 0.595 & 0.546 & 0.597 & 0.595 & 0.561 & 0.593 & 0.563 & 0.500 \\
WDP & \second{0.099} & \second{0.098} & \second{0.098} & \second{0.088} & \second{0.098} & \second{0.094} & \second{0.075} & \second{0.091} & \second{0.074} & \textbf{0.063} \\
CLIP2Protect & 0.147 & 0.147 & 0.147 & 0.144 & 0.147 & 0.148 & 0.148 & 0.148 & 0.149 & 0.140 \\
DiffAM & 0.200 & 0.200 & 0.200 & 0.193 & 0.200 & 0.198 & 0.178 & 0.195 & 0.178 & 0.159 \\
TIP-IM & 0.235 & 0.235 & 0.237 & 0.260 & 0.234 & 0.232 & 0.253 & 0.240 & 0.273 & 0.297 \\
DiffProtect & 0.339 & 0.339 & 0.338 & 0.310 & 0.339 & 0.338 & 0.326 & 0.338 & 0.327 & 0.300 \\
PGD & 0.188 & 0.189 & 0.190 & 0.227 & 0.182 & 0.188 & 0.247 & 0.203 & 0.270 & 0.306 \\
MI-FGSM & 0.169 & 0.171 & 0.170 & 0.191 & 0.165 & 0.168 & 0.212 & 0.178 & 0.229 & 0.274 \\
TI-DIM & 0.207 & 0.207 & 0.208 & 0.208 & 0.205 & 0.198 & 0.158 & 0.193 & 0.167 & 0.162 \\
Ours & \textbf{0.024} & \textbf{0.023} & \textbf{0.025} & \textbf{0.065} & \textbf{0.022} & \textbf{0.019} & \textbf{0.038} & \textbf{0.024} & \textbf{0.055} & \second{0.098} \\
\bottomrule
\end{tabular}
\end{table*}

\section{Adaptivity of SVD Truncation Bases}
\label{sec:supp-svd-freeze}

We investigate the adaptivity of SVD truncation through three complementary analyses. First, we quantify how the selected singular subspaces
vary across source images. Second, we track how these subspaces evolve during
optimization of the same image. Finally, we assess the effect of continued basis updates by comparing fully adaptive truncation with variants
whose bases are frozen at different optimization steps, measuring both protection performance and image fidelity.

\paragraph{Instance-dependent singular subspaces.}
We first analyze the subspaces selected by the full-resolution
DUNE branch at the final optimization step, using 500 CelebA-HQ
source images. For each RGB channel and held-out recognizer
configuration, we extract the leading rank-$r$ left and right
singular bases immediately before SVD truncation, with $r=11$.
For two orthonormal bases $Q_a,Q_b\in\mathbb{R}^{256\times r}$,
we measure subspace similarity as
\begin{equation}
S(Q_a,Q_b)
=\frac{\|Q_a^\top Q_b\|_F^2}{r}.
\label{eq:supp-subspace-similarity}
\end{equation}
This measure is invariant to singular-vector signs, ordering,
and orthogonal changes of basis within either subspace.
Identical subspaces give $S=1$, whereas independent uniformly
random rank-$r$ subspaces have expected similarity
$r/256=11/256\approx0.043$.

\begin{figure*}[t]
\centering
\includegraphics[width=0.95\textwidth]
  {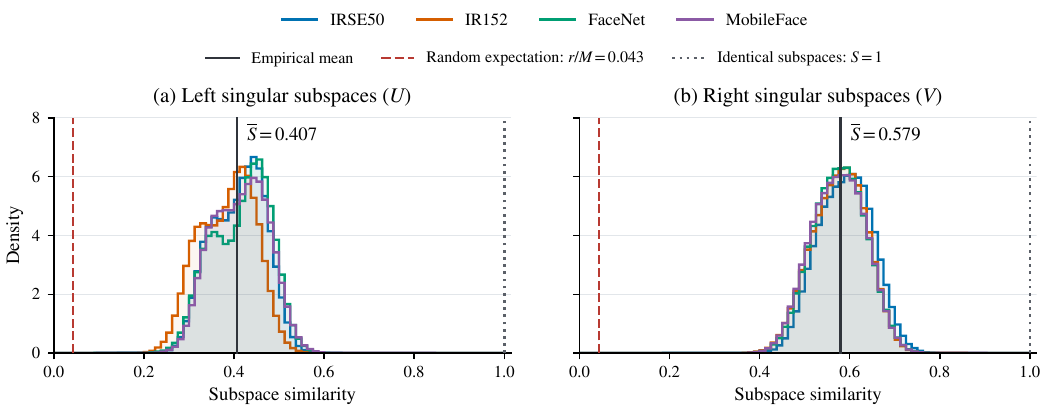}
\caption{
\textbf{Instance-dependent singular subspaces.}
Leading rank-11 left ($U$) and right ($V$) singular subspaces
of the full-resolution DUNE branch immediately before the
final SVD truncation, evaluated on 500 CelebA-HQ sources.
Each colored curve pools within-channel image-pair
similarities across RGB channels for one held-out
recognizer configuration.
The black vertical line marks the mean pooled over
all four configurations.
Reference lines indicate the expected similarity of
independent uniformly random rank-11 subspaces
($11/256$) and identical subspaces ($S=1$).
Pairwise observations share images and are not
independent samples.
}
\label{fig:supp-svd-instance}
\end{figure*}

We compute similarities between distinct images within each
channel and recognizer configuration, excluding self-comparisons.
As shown in \Cref{fig:supp-svd-instance},
the selected subspaces differ across images while retaining
substantial common structure: their similarities lie below the
identical-subspace reference and above the random-subspace
reference. This analysis characterizes the subspaces selected
at the SVD stage, rather than those of the final perturbations
after masking and clipping.

\paragraph{Subspace evolution within an image.}
We next examine whether the selected subspaces change during
optimization for the same source image.
Using the SVD-only setting of \Cref{tab:supp-svd-timing},
we record the leading rank-11 singular bases at steps
$10,30,50,70,$ and $100$ for 500 CelebA-HQ sources.
The bases are taken from the budget-clipped input to SVD
at each recorded update, before rank truncation.

For each anchor step $N\in\{10,30,50,70\}$,
we compare its selected subspace with those selected at
later recorded steps $t\ge N$, using
\Cref{eq:supp-subspace-similarity}.
For $Q\in\{U,V\}$, the mean similarity is
\begin{equation}
T_Q(t;N)
=
\frac{1}{n}\sum_{i=1}^{n}
S\!\left(Q_{i,r}^{(N)},Q_{i,r}^{(t)}\right),
\qquad t\ge N,
\label{eq:supp-temporal-subspace}
\end{equation}
where $n=500$ is the number of source images; channel and recognizer indices are omitted for clarity.
For the overall summary, similarities are averaged across
RGB channels and held-out configurations within each image
before computing the image mean and interquartile range.

\begin{figure*}[t]
\centering
\includegraphics[width=0.95\textwidth]
  {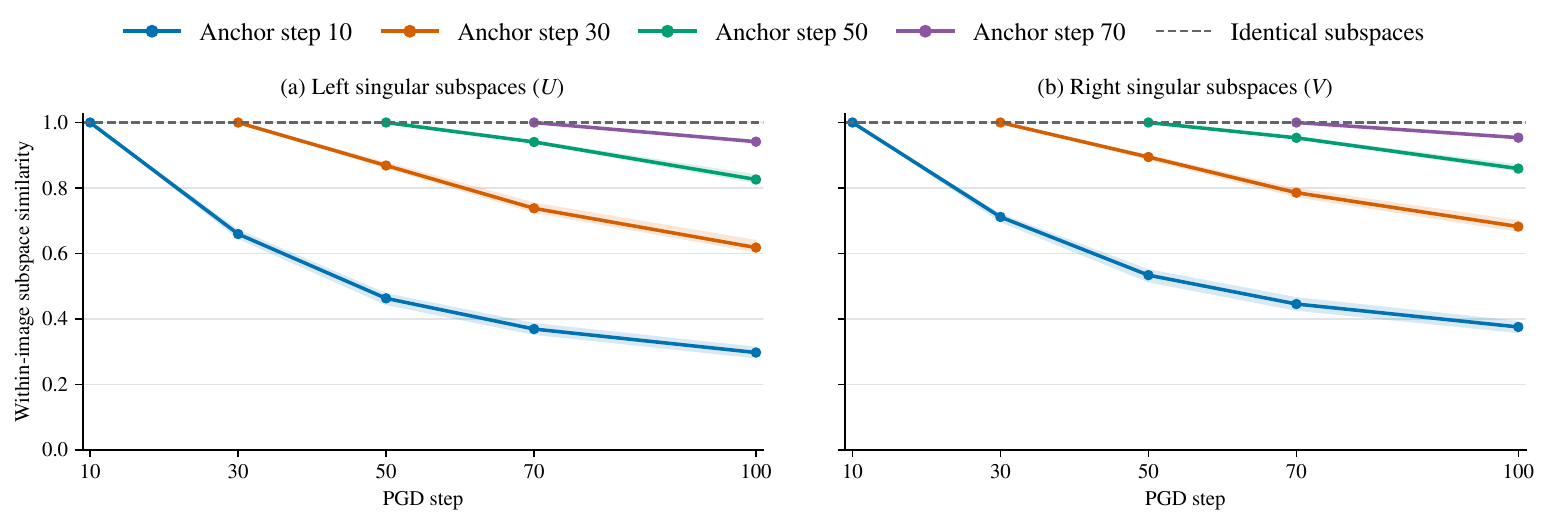}
\caption{
\textbf{Evolution of singular subspaces during optimization.}
Within-image similarities between the subspaces selected at
anchor steps $N\in\{10,30,50,70\}$ and later recorded steps.
Results use 500 CelebA-HQ sources under the SVD-only,
rank-11 protocol.
RGB channels and held-out configurations are averaged
within each image.
Lines show image means, and shaded bands indicate the
25th--75th percentiles across images, not confidence intervals.
Markers denote recorded steps.
}
\label{fig:supp-svd-temporal}
\end{figure*}

\Cref{fig:supp-svd-temporal} measures how closely the subspaces
selected later remain aligned with those selected at each
anchor step. A value of one indicates identical subspaces,
while smaller values indicate a change in the selected span.

At step 100, the mean similarities to the step-10 subspaces
are 0.297 for $U$ and 0.375 for $V$.
The corresponding similarities to the step-70 subspaces
are 0.941 and 0.954, respectively.
Thus, the selected subspaces change substantially within
the same image during optimization, while later anchors
remain more closely aligned with the final subspaces.

\paragraph{Effect of freezing the truncation bases.}
We examine whether the truncation basis must be updated throughout optimization. All variants run for 100 optimization steps; only the point at which the basis is frozen differs. Under the setting of \Cref{tab:supp-svd-timing}, the top-11 left and right singular vectors $U$ and $V$ of the perturbation are recomputed at every step until step $N$ and then fixed. The perturbation continues to be optimized for the remaining $100-N$ steps within the fixed row and column spaces. From step $N{+}1$ onward, each update is projected as $\delta \leftarrow U_N U_N^\top \delta\, V_N V_N^\top$, so only the $11\times11$ core matrix $C = U_N^\top \delta V_N$ is optimized while the column and row spaces remain fixed (\Cref{tab:supp-svd-freeze}). The adaptive column recomputes the basis at all 100 steps and corresponds to per-step SVD in \Cref{tab:supp-svd-timing}.

\begin{table}[!htbp]
\centering\footnotesize
\caption{\textbf{Adaptive versus frozen truncation bases (SVD only, rank 11) on CelebA-HQ.} All variants use 100 optimization steps. Adaptive recomputes the basis at every step; frozen variants fix it after step $N$. Dodging PSR (\%) averages four held-out recognizers.}
\label{tab:supp-svd-freeze}
\setlength{\tabcolsep}{4pt}
\resizebox{\columnwidth}{!}{%
\begin{tabular}{lccccc}
\toprule
 & Adaptive & \multicolumn{4}{c}{Frozen after step $N$} \\
\cmidrule(lr){2-2}\cmidrule(l){3-6}
Metric & All 100 steps & 70 & 50 & 30 & 10 \\
\midrule
PSR$\uparrow$ & 85.50 & 84.70 & 82.55 & 77.45 & 33.20 \\
PSNR$\uparrow$ & 31.47 & 31.78 & 32.19 & 32.91 & 34.07 \\
SSIM$\uparrow$ & 0.854 & 0.858 & 0.864 & 0.872 & 0.875 \\
FID$\downarrow$ & 45.3 & 43.6 & 41.3 & 37.8 & 31.0 \\
\bottomrule
\end{tabular}}
\end{table}

Freezing the basis confines all subsequent updates to a fixed subspace, so directions that become useful later cannot be added; PSR drops sharply when the basis is frozen early (33.20\% after step 10). Because DUNE recomputes the basis at every step, the subspace follows the perturbation throughout optimization and achieves the highest PSR. Fixed frequency filters such as DCT and FFT low-pass filtering (\Cref{tab:refinement}) use fixed, image-independent transform bases, whereas SVD truncation selects the basis from the current perturbation at every step. Together, these analyses show that SVD selects
image-dependent subspaces that evolve during optimization.
Under the evaluated protocol, updating these subspaces
throughout optimization yields higher dodging PSR than
freezing them, with a corresponding loss of image fidelity.
Comparisons with fixed frequency filters are reported
separately in \Cref{tab:refinement}.

\section{Robustness to Transformations}
\begin{figure}[!htbp]
\centering
\includegraphics[width=\linewidth]{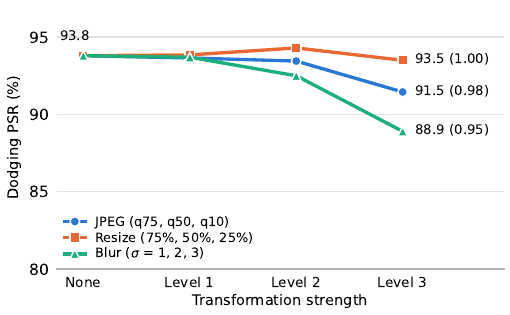}
\caption{\textbf{Transformation robustness on CelebA-HQ.} Dodging PSR after JPEG compression, resizing, and Gaussian blur, each at three strengths.}
\label{fig:robust-transforms}
\end{figure}

To measure how well the perturbation preserves identity disruption under common image transformations, we apply JPEG compression, resizing, and Gaussian blur to the protected images at three levels of increasing strength. The settings of each level are shown in \Cref{fig:robust-transforms}: JPEG quality 75, 50, and 10; resizing to 75\%, 50\%, and 25\%; and Gaussian blur with $\sigma=1$, 2, and 3. Protection is largely maintained even under strong transformations: at Level~3, dodging PSR remains 93.5\% for resizing, 91.5\% for JPEG, and 88.9\% for blur, retaining approximately 95\% or more of the untransformed PSR (93.8\%). Under the same transformations, DUNE consistently maintains strong identity disruption, keeping the identity similarity low at every level (\Cref{tab:supp-transform-cos}). The observed robustness is consistent with the structured perturbations encouraged by SVDG refinement and the bilinearly upsampled auxiliary branch.
\label{sec:transform-robustness}

We also purify the protected images with SDEdit using Stable Diffusion 2-base at strengths 0.1, 0.2, and 0.3, and measure the mean cosine similarity between each purified image and the dodging reference, a different portrait of the same identity, over the four held-out recognizers (\Cref{tab:supp-purify-cos}). Lower values indicate stronger identity disruption. Most of makeup- and diffusion-based methods retain lower similarity after purification. These methods alter the image with a generative model, so their changes remain natural facial content that the diffusion prior of the purifier largely preserves. In contrast, DUNE adds a perturbation that does not follow natural image statistics, which the purifier treats as noise and regenerates. Nevertheless, DUNE maintains substantially lower similarity than the noise-based methods at every strength.
\begin{table}[H]
\centering\footnotesize
\caption{\textbf{Identity similarity after diffusion purification on CelebA-HQ.} Mean cosine similarity between each query image and the dodging reference, a different unmodified portrait of the same identity, over four held-out recognizers (lower is better); Clean denotes the unprotected source images. A larger SDEdit strength $s$ indicates stronger purification.}
\label{tab:supp-purify-cos}
\setlength{\tabcolsep}{5pt}
\begin{tabular}{lcccc}
\toprule
Method & No purification & $s=0.1$ & $s=0.2$ & $s=0.3$ \\
\midrule
Clean & 0.597 & 0.548 & 0.497 & 0.445 \\
WDP & 0.099 & \textbf{0.134} & \textbf{0.159} & 0.171 \\
CLIP2Protect & 0.147 & 0.158 & 0.163 & \textbf{0.162} \\
DiffAM & 0.200 & 0.206 & 0.208 & 0.206 \\
TIP-IM & 0.235 & 0.283 & 0.303 & 0.299 \\
DiffProtect & 0.339 & 0.310 & 0.287 & 0.263 \\
PGD & 0.188 & 0.299 & 0.319 & 0.306 \\
MI-FGSM & 0.169 & 0.238 & 0.268 & 0.274 \\
TI-DIM & 0.207 & 0.236 & 0.262 & 0.268 \\
Ours & \textbf{0.024} & 0.143 & 0.199 & 0.221 \\
\bottomrule
\end{tabular}
\end{table}

\section{Generation Cost}
\label{sec:supp-cost}

\Cref{tab:cost} compares the time and GPU memory required to generate one protected image. Because DUNE directly optimizes the perturbation, it avoids costly generator training and diffusion inversion.

\begin{table}[H]
\centering\small
\caption{\textbf{Generation cost per protected image} on one RTX 3090. Training is a one-time cost; ``--'' denotes none.}
\label{tab:cost}
\footnotesize\setlength{\tabcolsep}{4pt}
\begin{tabular}{lrrr}
\toprule
Method & s/image & GPU (MiB) & Training (h) \\
\midrule
AMT-GAN & 0.20 & 2438 & 15.93 \\
DiffAM (editing) & 0.73 & 6072 & 1.76 \\
WDP & 26.23 & 17254 & -- \\
TIP-IM & 29.74 & 1660 & -- \\
DiffProtect & 38.47 & 14458 & -- \\
CLIP2Protect & 47.63 & 7974 & -- \\
\midrule
MI-FGSM & 1.05 & 2544 & -- \\
PGD & 1.11 & 2544 & -- \\
TI-DIM & 1.37 & 2544 & -- \\
\midrule
\textbf{Ours} & 3.67 & 2706 & -- \\
\bottomrule
\end{tabular}
\end{table}

Diffusion-based and makeup-based methods either require separate generator training or take a long time to generate each image, and many of them rely on heavy models that consume large amounts of GPU memory. In contrast, DUNE requires no training, takes 3.67\,s per image, and uses 2706\,MiB of memory, close to the simple noise attacks. DUNE thus achieves the strong protection reported in \Cref{tab:faceswap} at a moderate generation cost.

\section{Optimization Algorithm}
\label{sec:supp-algorithm}
\Cref{alg:dualres} follows the objective and updates in
main-paper \Cref{sec:formulation,sec:opt}.
Here $\mathcal R_{\mathrm{SVDG}}$ denotes one application of
the SVDG block, including the branch clipping operations
specified in \Cref{sec:opt}.
Both branches are initialized with unrefined uniform noise.
The SVDG block is applied exactly once after the main-branch
gradient update in each iteration, for $T$ applications in total.

\begin{algorithm}[H]
\caption{Dual-resolution face protection}
\label{alg:dualres}
\small
\begin{algorithmic}[1]
\Require Portrait $x$, surrogate recognizers $\{f_k\}_{k=1}^K$
\Require Grid size $(h,w)$, $T,\alpha,\epsilon,\epsilon_r,\epsilon_g,r,\sigma_r,\lambda$
\Ensure Protected source $x_{\mathrm p}$
\State $S\gets$ smoothed gradient-importance map from 20 probes (\Cref{sec:mtv})
\State $k\gets\lfloor0.3HW\rfloor$
\State $\tau\gets$ the $k$-th largest score in $S$
\State $M_{ij}\gets\mathbf1[S_{ij}\ge\tau]$ \Comment{Top $30\%$; fixed}
\State $\dlow\sim\mathcal U(-\epsilon_g,\epsilon_g)^{3\times h\times w}$
\State $\dhigh\sim\mathcal U(-\epsilon_r,\epsilon_r)^{3\times H\times W}$
\For{$t=0,\ldots,T-1$}
    \State $v\gets\dhigh+U(\dlow)$ \Comment{Sum branch outputs}
    \State $\delta\gets C_{-\epsilon,\epsilon}(M\odot v)$ \Comment{One shared mask}
    \State $x_{\mathrm p}\gets C_{0,1}(x+\delta)$
    \State $\eta\gets x_{\mathrm p}-x$; evaluate \Cref{eq:loss}
    \State $(g_r,g_g)\gets(\nabla_{\dhigh}\mathcal L,\nabla_{\dlow}\mathcal L)$
    \Statex \(\triangleright\) \textit{Main SVD branch: normalized gradient}
    \State $\widehat g_r\gets$ normalized direction in \Cref{eq:grad-direction}
    \State $\dhigh\gets\mathcal R_{\mathrm{SVDG}}(\dhigh-\alpha\,\widehat g_r)$ \Comment{Once per iteration}
    \Statex \(\triangleright\) \textit{Auxiliary grid: sign-PGD}
    \State $\dlow\gets C_{-\epsilon_g,\epsilon_g}(\dlow-\alpha\,\mathrm{sign}(g_g))$
\EndFor
\State \Return $C_{0,1}(x+C_{-\epsilon,\epsilon}(M\odot(\dhigh+U(\dlow))))$
\end{algorithmic}
\end{algorithm}

\FloatBarrier

\end{document}